\documentclass[11pt,a4paper]{article}

\usepackage{hyperref}
  \usepackage{xcolor}
 \definecolor{darkblue}{rgb}{0, 0, 0.5}
 \hypersetup{colorlinks=true,citecolor=darkblue, linkcolor=darkblue, urlcolor=darkblue}

\usepackage[acceptedWithA]{tacl2018v2}
\usepackage{times,latexsym}
\usepackage{url}
\usepackage[T1]{fontenc}
\definecolor{amber}{rgb}{1.0, 0.49, 0.0}
\definecolor{cadmiumgreen}{rgb}{0.0, 0.42, 0.24}
\definecolor{palechestnut}{rgb}{0.87, 0.68, 0.69}
\definecolor{lavender(web)}{rgb}{0.9, 0.9, 0.98}
\definecolor{lemonchiffon}{rgb}{1.0, 0.98, 0.8}
\definecolor{amber}{rgb}{1.0, 0.75, 0.0}
\definecolor{darkgoldenrod}{rgb}{0.72, 0.53, 0.04}
\definecolor{darkbrown}{rgb}{0.4, 0.26, 0.13}

\usepackage{xspace,mfirstuc,tabulary}

\newif\iftaclinstructions
\taclinstructionsfalse 
\iftaclinstructions
\renewcommand{\confidential}{}
\renewcommand{\anonsubtext}{(No author info supplied here, for consistency with
TACL-submission anonymization requirements)}
\newcommand{\instr}
\fi

\iftaclpubformat 

\else

\fi

\usepackage{verbatimbox}
\usepackage{natbib}
\usepackage{graphicx}
\usepackage{fancyvrb}
\usepackage{booktabs}
\usepackage{comment}
\usepackage{amssymb}
\usepackage{pifont}
\usepackage{adjustbox}
\usepackage[makeroom]{cancel}
\usepackage{mathtools}
\usepackage{wasysym}
\usepackage{tipa}

\newcommand\notsotiny{\@setfontsize\notsotiny{6}{7}}

\newcommand{\itriple}[3]{$\langle #1$, #2, $\text{#3}\rangle$}
\newcommand{\rtriple}[3]{$\langle #1$, #2, $#3\rangle$}

\usepackage{listings}
\usepackage{subcaption}

\usepackage{gb4e}

\usepackage[11pt]{moresize}
\usepackage{amsmath}
\usepackage{amsthm}
\usepackage[normalem]{ulem}
\usepackage{bm}

\newcommand{\norm}[1]{\left\lVert#1\right\rVert}

\newcommand{\stkout}[1]{\ifmmode\text{\sout{\ensuremath{#1}}}\else\sout{#1}\fi}
\DeclareMathOperator{\expcust}{exp}

\newcommand{\cmark}{\textcolor{cadmiumgreen}{\ding{51}}}%
\newcommand{\xmark}{\textcolor{red}{\ding{55}}}

\newcommand{\bsqrt}{\text{\color{darkbrown}{$\pmb{\sqrt{~}}$}\color{black}}}

\usepackage{textcomp}

\noautomath
\title{AMR Similarity Metrics from Principles}
  
  \author{Juri Opitz {\normalfont and} Letitia Parcalabescu {\normalfont and} Anette Frank \\
  Department for Computational Linguistics \\
  Heidelberg University \\
  69120 Heidelberg \\
 {\tt $\lbrace$opitz,parcalabescu,frank$\rbrace$@cl.uni-heidelberg.de} }
\date{June 2019}

\begin{document}

\maketitle

\begin{abstract}
Different metrics have been proposed to compare \textit{Abstract Meaning Representation (AMR)} graphs. The canonical \textsc{Smatch} metric \cite{cai-knight-2013-smatch} aligns the variables of two graphs and assesses
triple matches. The recent \textsc{SemBleu} metric \cite{DBLP:journals/corr/abs-1905-10726} is based on the machine-translation metric \textsc{Bleu} \cite{papineni2002bleu} and increases computational efficiency by ablating the variable-alignment.

In this paper, i) we establish criteria that enable researchers to perform a \textit{principled assessment of metrics} comparing meaning representations like AMR; ii) we undertake a \textit{tho\-rough a\-na\-ly\-sis} of \textsc{Smatch} and \textsc{SemBleu} where we show that the latter exhibits some undesirable properties. For example, it does not conform to the \textit{iden\-ti\-ty of indiscernibles} rule and introduces biases that are hard to control; 
iii) we propose a \textit{no\-vel metric \textsc{S$^2$match}} that is more benevolent to only very slight meaning de\-vi\-at\-ions and targets the fulfilment of all estab\-lis\-hed criteria. We assess its suitability 
and show its advantages over \textsc{Smatch} and \textsc{SemBleu}.

\end{abstract}

\section{Introduction}
\begin{figure}
    \centering
    \includegraphics[width=0.9\linewidth]{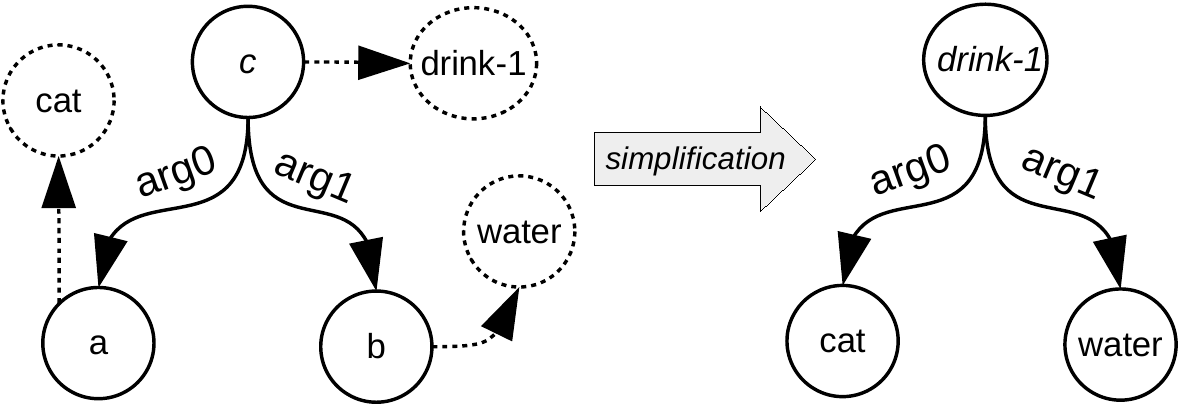}
    \caption{\textit{A cat drinks water.} Simplified AMR graph 
    and underlying deep form with \textit{is-instance} relations (\protect\includegraphics[scale=0.5]{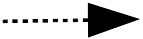}) from variables (solid) to concepts (dashed).}
    \label{fig:simplereduction}
    \vspace{-3mm}
\end{figure}

Proposed in 2013, the aim of Abstract Meaning Representation (AMR) is to represent a sentence's meaning in a machine-readable graph format \cite{banarescu2013abstract}.
AMR graphs are rooted, acyclic, directed and edge-labeled.
Entities, events, properties and states are represented as \textit{variables} that are linked to corresponding \textit{concepts} (encoded as leaf nodes) via \textit{is-instance} relations (cf.\ Figure \ref{fig:simplereduction}, left). This structure allows us to capture complex linguistic phenomena such as coreference, semantic roles or polarity.

When measuring the similarity between two AMR graphs $A$ and $B$, for instance for the purpose of AMR parse quality evaluation, the metric of choice is usually \textsc{Smatch} \cite{cai-knight-2013-smatch}. Its backbone is an alignment-search between the graphs' variables. Recently, the \textsc{SemBleu} metric  \cite{DBLP:journals/corr/abs-1905-10726} has been proposed that operates on the basis of a variable-free AMR (Figure \ref{fig:simplereduction}, right)\footnote{Most research papers on AMR display the graphs in this `shallow' form. This increases simplicity and readability.  
\cite{lyu-titov-2018-amr,konstas2017neural,DBLP:journals/corr/zhang19,DBLP:journals/corr/damonte19,DBLP:journals/corr/SongZPWG16}.}, converting it to a bag of k-grams. Circumventing a variable alignment search re\-du\-ces computational cost and ensures full determinacy. Also, grounding the metric in \textsc{Bleu} \cite{papineni2002bleu} has a certain appeal, since \textsc{Bleu} is quite popular in Machine Translation.  

However, we find that we are lacking a principled in-depth comparison of the properties of different AMR metrics which would help informing researchers to answer questions such as: \textit{Which metric should I use to assess the similarity of two AMR graphs, e.g., in AMR parser evaluation? What are the trade-offs when choosing
one metric over the other?} Besides providing criteria for such a principled comparison, we  discuss a property that none of the existing AMR metrics currently satisfies: they do not measure graded meaning differences. Such differences may emerge due to near-synonyms such as \textit{ruin -- annihilate; skinny -- thin -- slim; 
enemy -- foe} \cite{Inkpen:2006:BUL:1169207.1169210,Edmonds:2002:NLC:643081.643082} or   paraphrases such as \textit{be able to -- can; unclear -- not clear}.  
In a classical syntactic parsing task, metrics do not  need to address this issue since input tokens  are typically projected to lexical concepts by lemmatization, hence two graphs for the same sentence tend not to disagree on the concepts  projected from the input. This is different in semantic parsing where the projected concepts are often more abstract.

This article
is structured as follows: We first establish \textit{seven principles} that one may expect a metric for comparing meaning representations to satisfy,
in order to obtain meaningful and appropriate scores  for the given purpose (\S \ref{sec:principles}). Based on these principles we provide an \textit{in-depth analysis} of the properties of the AMR metrics \textsc{Smatch} and \textsc{SemBleu} (\S \ref{sec:metricanalysis}). We then \textit{develop \textsc{S$^2$match}, an extension of \textsc{Smatch}} that abstracts away from a purely symbolic level, allowing for a graded semantic comparison of atomic graph-elements (\S \ref{sec:s2}). By this move, we enable \textsc{Smatch} to take into account fine-grained meaning differences. We show that our proposed metric retains
valuable benefits of \textsc{Smatch}, but at the same time is more benevolent to slight meaning deviations. Our code is available online \url{https://github.com/Heidelberg-NLP/amr-metric-suite}. 

\section{From principles to AMR metrics}
\label{sec:principles}

The problem of comparing AMR graphs  $A,B \in \mathcal{D}$ with respect to the meaning they express occurs in several scenarios, for example, parser evaluation or inter-annotator agreement calculation (IAA). To measure the extent to which $A$ and $B$ agree with each other, we need a $metric$: $\mathcal{D} \times \mathcal{D} \rightarrow \mathbb{R}$ that returns a \textit{score} reflecting \textit{meaning distance} or \textit{meaning similarity} (for convenience, we use similarity). Below we establish seven principles that seem desirable for this metric.

\subsection{Seven metric principles}

 The first four metric principles are \textbf{mathematically motivated}:

\textit{\textbf{I.\  continuity, non-negativity and upper-bound}} A similarity function should be continuous, with two natural edge cases: $A,B$ are equivalent (maximum similarity) or  unrelated (minimum similarity). 
By choosing 1 as upper bound, we obtain 
the following constraint on  $metric$: $\mathcal{D} \times \mathcal{D} \rightarrow [0,1]$.\footnote{At some places in this paper, due to conventions, we project this score onto [0,100] and speak of \textit{points}.}

\textit{\textbf{II.\  identity of indiscernibles}} ~This focal principle is formalized by 
$metric(A,B) = 1 \Leftrightarrow A=B$. It is violated if a metric assigns a value indicating equivalence 
to inputs that are not equivalent or if it considers equivalent inputs as different. 
 
\textit{\textbf{III.\  symmetry}} In many cases, we want a metric to be sym\-met\-ric:  $metric(A,B)$ = $metric(B,A)$. A met\-ric violates this principle if it assigns a pair of ob\-jects different scores when  argument order is in\-ver\-ted. 
Together with principles I and II, it 
extends the scope of the metric 
to usages beyond
parser evaluation, as
it also enables sound IAA calculation, clustering 
and classification of AMR graphs when we use the metric as a kernel (e.g., SVM). In parser evaluation, one may dispense with any (strong) requirements for symmetry---however, the metric must 
then 
be applied in a standardized way, with a fixed order of arguments. 

In cases where there is no defined reference, the asymmetry could be handled by aggregating $metric(A,B)$ and $metric(B,A)$, e.g., using the mean. However, it is open what aggregation is best suited and how to interpret results, e.g. for $metric(A,B)=0.1$ and $metric(B,A)=0.9$.

\textit{\textbf{IV.\  determinacy}} Repeated calculation over the same inputs should yield
the same score. This principle is clearly desirable as it ensures re\-pro\-du\-ci\-bi\-lity (a very small deviation may be tolerable).

The 
next three principles we believe to be desirable specifically when comparing meaning representation graphs such as AMR \cite{banarescu2013abstract}. 
The first two of the following principles are \textbf{motivated by computer science and linguistics}, whereas the last one is \textbf{motivated from a linguistic and an engineering perspective}.

\textit{\textbf{V.\  no bias}}: Meaning representations consist of 
nodes and edges encoding
specific information types. 
Un\-less 
explicitly justified, a 
metric should not un\-jus\-ti\-fi\-ably or in unintended ways favor cor\-rect\-ness or penalize errors for specific sub\-struc\-tures (e.g., 
leaf nodes). In case a metric favors or penalizes certain substructures more than others, in the interest of transparency, this should be made clear and explicit, and should be easily verifiable and consistent. E.g., if we wish to give negation of the main predicate of a sentence a two times higher weight compared to ne\-ga\-tion in an embedded sentence, 
we want this to be 
made 
transparent. 
A concrete example for a transparent bias is found in \citet{cai-lam-2019-core}. They analyze the impact of their novel top-down AMR parsing strategy by
integrating a root-distance bias into \textsc{Smatch} 
to focus on structures situated at the top of a graph.

We now turn to properties that focus on the nature of the objects we 
aim
to compare: graph-based compositional meaning representations. These graphs consist of ato\-mic conditions that determine the circumstances under which a sentence is true. Hence, our $metric$ score should increase with increasing overlap of $A$ and $B$, which we denote 
$f(A,B)$, the number of \textit{matching} conditions. This overlap can be viewed from a \textbf{symbolic} or/and a \textbf{graded} perspective (cf., e.g., \citet{10.5555/1121730} who denote these perspectives as `syntactic' vs.\ `semantic'). From the symbolic perspective, we compare the nodes and edges of two graphs on a symbolic level, while from the graded perspective, we take into account the degree to which nodes and edges differ. 
Both types of matching involve a precondition
: If $A$ and $B$ contain variables, we need a variable-mapping in order to match conditions from $A$ and $B$.\footnote{E.g., consider graph $A$ in Figure \ref{fig:simplereduction} and its  set of triples $t(A)$:
\{\itriple{x_1}{instance}{drink-1} \itriple{x_2}{instance}{cat},  \rtriple{x_1}{arg0}{x_2}, \rtriple{x_1}{arg1}{x_3},  \itriple{x_3}{instance}{water}\}. When comparing  $A$ against graph $B$ we need to judge whether a triple $\mathbf{t} \in t(A)$
is also contained in $B$: $\mathbf{t} \in t(B)$. For this, we need a mapping $map$: $vars(A)$ $\rightarrow$ $vars(B)$ where $vars(A)$ $=$ $\{x_1,..,x_n\}$,  $vars(B)$ = $\{y_1,..,y_m\}$ s.t.\ $f$ is maximized.}

\textit{\textbf{VI.\  matching (graph-based) meaning representations -- symbolic match
}}\label{princ:fol}  
A natural symbolic overlap-objective can be found in the Jaccard index $J$ \cite{jaccard1912distribution,real1996probabilistic,papadimitriou2010web}: Let $t(G)$ be the set of triples of graph $G$, $f(A,B) = |t(A)\cap t(B)|$ the size of the overlap of $A,B$, and $z(A,B)$ = $|t(A) \cup t(B)|$ the size of their union. Then, we wish that $A$ and $B$ are considered more similar to each other than $A$ and $C$  iff $A$ and $B$ exhibit a greater relative agreement in their (symbolic) conditions:
$metric(A,B)$ $>$ $metric (A,C)$ $\Leftrightarrow$ $ \frac{f(A,B)}{z(A,B)} = J(A,B) > \frac{f(A,C)}{z(A,C)} = J(A,C)$. An allowed exception to this monotonic relationship can occur if we want to take into account a graded semantic match of atomic graph elements or sub-structures, which we will now elaborate on. 

\textit{\textbf{VII.\  matching (graph-based) meaning representations -- graded semantic match}}:  One motivation for this principle can be found in engineering, e.g., when assessing the quality of produced parts. Here, small deviations from a reference may be tolerable within certain limits. 
Similarly, two AMR graphs may match almost perfectly -- except for two small divergent components. The extent of divergence can be measured by the degree of similarity of the two divergent components. In our case, we need linguistic know\-led\-ge to judge 
what degree of divergence we are dealing with and whether it is tolerable.

For example, consider that graph $A$ contains a triple $\langle x,\text{instance}, \text{\textit{conceptA}}\rangle$ and graph $B$ a triple
$\langle y,\text{instance}, \text{\textit{conceptB}}\rangle$, while otherwise the graphs are equivalent, and the alignment has set $x$$=$$y$. 
Then $f(A,B)$ should be higher when \textit{conceptA} is similar to \textit{conceptB} compared 
to
the case where \textit{conceptA} is dissimilar to \textit{conceptB}. In AMR, concepts are often abstract, so near-\-sy\-no\-nyms may even be fully admissible (\textit{enemy--foe}). 
While such (near-)synonyms are bound to occur frequently when we compare AMR graphs of \textit{different sentences} that may contain paraphrases, we will 
see, in Section \S4, that this can also occur in parser evaluation, where two different graphs represent the \textit{same sentence}.  By defining $metric$ to map to a range  [0,1] we already defined it to be globally graded. 
Here, we desire that \textit{graded similarity} 
may also
hold of \textit{minimal units} of AMR graphs, such as atomic concepts or even 
sub-graphs, e.g., to reflect that $injustice(x)$ is very similar to $justice(x) \land polarity(x,-)$.

\subsection{AMR metrics: \textsc{Smatch}  and \textsc{SemBleu}}

With our seven principles for AMR similarity metrics in place, we now introduce \textsc{Smatch} and \textsc{Sem\-Bleu}, two metrics that differ in their design and assumptions. We describe each of them in detail 
and summarize their differences, setting the stage for our in-\-dep\-th metric analysis (\S \ref{sec:metricanalysis}).

\paragraph{Align and match -- \textsc{Smatch}}  The  \textsc{Smatch} metric 
operates in two steps. First, (i) we align the variables in $A$ and $B$ in the best possible way, by finding a mapping
$map^\star$: $vars(A)$ $\rightarrow$ $vars(B)$ that 
yields a maximal set of matching triples between $A$ and $B$. E.g., if \rtriple{x_i}{rel}{x_j} $\in$ $t(A)$ and \rtriple{map^\star(x_i)}{rel}{map^\star(x_j)} =  \rtriple{y_k}{rel}{y_m} $\in$ $t(B)$, we obtain one triple match. (ii) We compute Precision, Recall and F1 score based on the set of triples returned by the alignment search. 
The NP-hard alignment search problem of step (i) is solved 
with a greedy hill-climber: Let $f_{map}(A,B)$ be the count of matching triples under any mapping function \textit{map}. Then,\\  

\vspace*{-7mm}
\begin{equation}
    map^\star = \operatorname*{argmax}_{map} f_{map}(A,B).
\end{equation}

Multiple restarts with different seeds increase the likelihood of finding better optima. 

\paragraph{Simplify and match -- \textsc{SemBleu}}\label{par:sembleudecript} The \textsc{SemBleu} metric in \citet{DBLP:journals/corr/abs-1905-10726} can also be described as a two-step procedure.
But unlike \textsc{Smatch} it operates on a \textbf{variable-free reduction} of an AMR graph $G$, which we denote by $G^{vf}$ ($vf$: variable-free, Figure \ref{fig:simplereduction}, right-hand side).

In a first step, (i) \textsc{SemBleu} performs k-gram ex\-trac\-tion  from $A^{vf}$ and $B^{vf}$ in a breadth-first tra\-ver\-sal (path extraction). It then (ii) adopts
the \textsc{Bleu} score from MT \cite{papineni2002bleu} to calculate an overlap score based on the extracted bags of k-grams:
\begin{align}\label{eq:sembleu}
    \textsc{SemBleu} &= BP \cdot \expcust \Bigg( \sum_{k=1}^{n} w_k \log p_k \Bigg) \\
    BP &= e^{\min \big\{ 1- \frac{|B^{vf}|}{|A^{vf}|}, 0\big\} }
\end{align}

where 
$p_k$ is \textsc{Bleu}'s  \textit{modified k-gram precision} that measures $k$-gram overlap of a candidate against a reference: 
$p_k$ = $\frac{|kgram(A^{vf}) \cap kgram(B^{vf})|}{|kgram(A^{vf})|}$.
$w_k$ is the (typically uniform) weight over chosen k-gram sizes. \textsc{Sembleu} uses NIST geometric probability smoothing \cite{chen-cherry:2014:W14-33}.
The recall-focused `brevity penalty' $BP$ returns a value smaller than 1 when the candidate length $|A^{vf}|$ is smaller than the reference length $|B^{vf}|$.

The graph traversal performed in \textsc{SemBleu} 
starts at the root node. During this traversal it simplifies the graph by replacing variables with their corresponding concepts (see Figure \ref{fig:simplereduction}: the node \textit{c} becomes \textsc{Drink-01}) and 
collects visited nodes and edges in uni-, bi- and tri-grams (k=3 is recommended). Here, a source node  together with a relation and its target node counts as a bi-gram.
For the graph in Figure \ref{fig:simplereduction}, the extracted unigrams are $\{cat,~water,~drink$-$01\}$;
the 
extracted bi-grams are $\{drink$-$01~arg1~cat,~drink$-$01~arg2~ water\}$.

\paragraph{\textsc{Smatch} vs.\ \textsc{SemBleu} in a nutshell} \textsc{SemBleu} 
differs significantly from \textsc{Smatch}.
A 
key
difference is 
that \textsc{SemBleu} 
operates on re\-du\-ced variable-free AMR graphs ($G^{vf}$) -- instead of full-fledged AMR graphs.  By eliminating variables, \textsc{SemBleu} bypasses an
 alignment search. This makes the calculation faster and alleviates a weakness of \textsc{Smatch}: the hill-climbing search is slightly imprecise. However, \textsc{SemBleu} is not guided by aligned variables as anchors. Instead, \textsc{SemBleu} uses an n-gram statistic (\textsc{Bleu}) to compute an overlap score for graphs, based on \textit{k-hop paths} extracted from $G^{vf}$, using the root node as the start for the extraction process. 
\textsc{Smatch}, by contrast, 
acts directly on variable-bound graphs  matching triples based on a selected alignment. If in some application we wanted it, 
both metrics allow the capturing of more `global' graph properties: \textsc{SemBleu} can increase its k-parameter and \textsc{Smatch} may match conjunctions of (interconnected) triples. In the following analysis, however, we will adhere to their default configurations since this is how they are used in most applications.

\section{Assessing AMR 
metrics with principles}\label{sec:metricanalysis} 

This section evaluates \textsc{Smatch} and \textsc{SemBleu} against the seven principles we estab\-lis\-hed above by asking:
\textit{Why does a metric satisfy or vio\-late a given principle?} and \textit{What does this imply?} 
We start with 
 principles  from mathematics.

 \label{subsec:laws}
 \label{par:noneg}\paragraph{I.\  Continuity, non-negativity and upper-bound} This principle is fulfilled by both metrics as they are functions of the form $metric: \mathcal{D} \times \mathcal{D} \rightarrow [0,1]$.

\noindent
\textbf{II.\  Identity of indiscernibles}\label{par:id} This principle is fun\-da\-men\-tal: An AMR metric must return ma\-xi\-mum score if and only if the graphs are equivalent in meaning. 
Yet, there are cases where \textsc{Sem\-Bleu}, in contrast to \textsc{Smatch}, 
does not satisfy this prin\-ci\-ple. Figure \ref{fig:idviolation} shows an example.

\begin{figure}[t]
\centering
\begin{ssmall}

\begin{Verbatim}[commandchars=\\\{\}]
\textbf{----------A------------Input-------------B----------}
\textbf{( p / predicate-01             ( p / predicate-01}
\textbf{ :ARG0 ( \colorbox{lavender(web)}{\textcolor{red}{x1} / man})             :ARG0 ( \colorbox{lavender(web)}{\textcolor{red}{x1} / man})}
\textbf{ :ARG1 ( \colorbox{lemonchiffon}{\textcolor{blue}{x2} / man})             :ARG1 \colorbox{lavender(web)}{\textcolor{red}{x1}} }
\textbf{ :ARG2 \colorbox{lemonchiffon}{\textcolor{blue}{x2}} )                    :ARG2 ( \colorbox{lemonchiffon}{\textcolor{blue}{x2} / man}))}
\textbf{-----------------------Scores-----------------------}
\textbf{\textsc{Smatch} ->  0.667} 
\textbf{\textsc{SemBleu}   ->  1.0}
\textbf{----------------------------------------------------}
\end{Verbatim}
\end{ssmall}

\caption{Two AMRs with semantic roles filled differently, \textsc{Sem\-Bleu} considers them as equivalent.}
\label{fig:idviolation}
\vspace{-3mm}
\end{figure}
Here, \textsc{Sem\-Bleu} yields a perfect score for 
two AMRs that differ in a single but 
crucial aspect: 
two of its \textsc{arg$_x$}  roles  are filled with arguments that are meant to refer to distinct individuals that share the same concept. The graph on the left is an ab\-stra\-ction of, e.g.\ \textit{The man$_{1}$ sees the other man$_2$ in the other man$_2$}, while the graph on the right 
is an abstraction of \textit{The man$_1$ sees himself$_1$ in the other man$_2$}. 
 \textsc{SemBleu} does not recognize
the dif\-fe\-ren\-ce in meaning between a reflexive and a non-\-re\-fle\-xive relation, assigning maximum si\-mi\-la\-ri\-ty score, whereas \textsc{Smatch} reflects such differences appropriately since it accounts for variables.

In sum, \textsc{SemBleu} does not satisfy
principle II because it operates on a var\-i\-ab\-le-free reduction of AMRs ($G^{vf}$). 
One could address this prob\-lem by reverting to canonical AMR graphs and 
adopting  variable alignment in \textsc{SemBleu}. But this would adversely affect the advertised efficiency advantages over \textsc{Smatch}. Re-integrating the alignment step would make \textsc{SemBleu} \textit{less} efficient than \textsc{Smatch} since it would add the complexity of breadth-first traversal, yielding a total complexity of $\mathcal{O(}$\textsc{Smatch}$\mathcal{)}$ plus $\mathcal{O}(|V| + |E|)$. 

\paragraph{III.\  Symmetry}\label{par:symm}

\begin{figure}

\begin{ssmall}
\begin{Verbatim}[commandchars=\\\{\}]
\textbf{----------\textcolor{red}{A}------------Input-------------\textcolor{blue}{B}----------}
\textbf{(a / and            (k7 / know-01}
\textbf{  :op1 (h / heat-01   :ARG0 (i / i}
\textbf{    :ARG1 (t / thing)   :ARG0-of (d9 / do-02}
\textbf{    :loc (b / between     :ARG1 t8}
 \textbf{     :op1 (w / we))      :ARG1 (t0 / thing}
\textbf{      :degree (s / so))     :ARG1-of (h2 / heat-01}
\textbf{    :op2 (k / know-01         :degree (s1 / so) }
\textbf{      :polarity -               :loc (b3 / between}
\textbf{      :ARG0 (i / i)             :op1 (w4 / we))))))}
\textbf{      :ARG1 (t2 / thing       :ARG1 (t8 / thing)}
\textbf{      :ARG1-of (d / do-02)))) :polarity -)}
\textbf{-----------------------Scores-----------------------}
\textbf{\textsc{SemBleu} (\textcolor{red}{A},\textcolor{blue}{B}) = 0.422 \colorbox{lemonchiffon}{\textcolor{red}{<<}} \textsc{SemBleu}} \textbf{(\textcolor{blue}{B},\textcolor{red}{A}) = 0.803)}
\textbf{\textsc{Smatch}  (\textcolor{red}{A},\textcolor{blue}{B}) = 0.829 \colorbox{lemonchiffon}{\textcolor{cadmiumgreen}{==}} \textsc{Smatch}}  \textbf{(\textcolor{blue}{B},\textcolor{red}{A}) = 0.829)}
\textbf{-----------------------------------------------------}
   \end{Verbatim}
   \end{ssmall}
\caption{Symmetry violation for two parses of \textit{Things are so heated between us, I don't know what to do}.}
\label{fig:violation-2}
\end{figure}

\begin{figure*}[ht]
\begin{subfigure}{.33\textwidth}
  \centering
  \includegraphics[width=\linewidth,trim=0 0.85cm 0 0.95cm,clip]{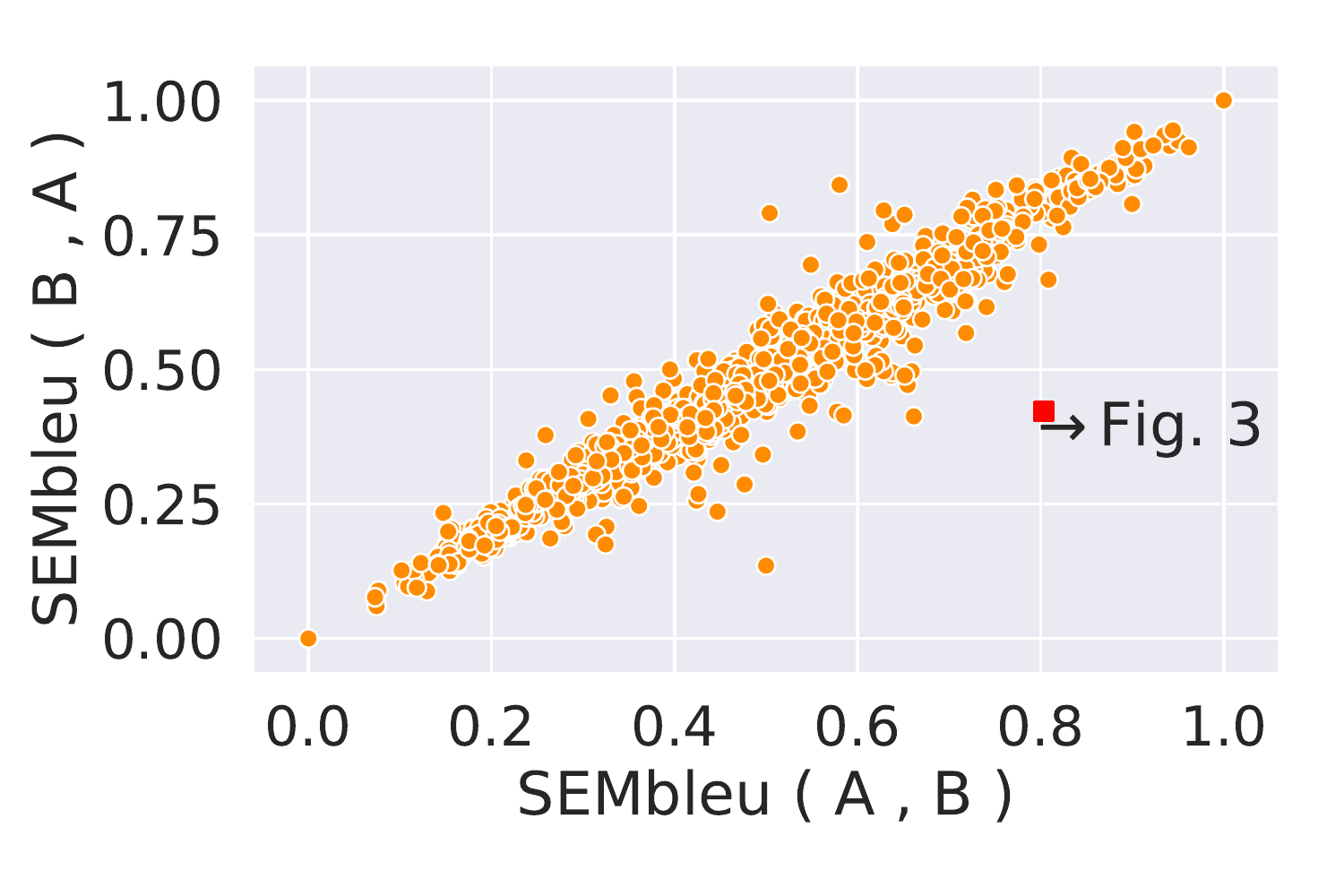}
  \caption{GPLA, Gold}
  \label{fig:sfig1}
\end{subfigure}%
\begin{subfigure}{.33\textwidth}
  \centering
  \includegraphics[width=\linewidth,trim=0 0.85cm 0 0.95cm,clip]{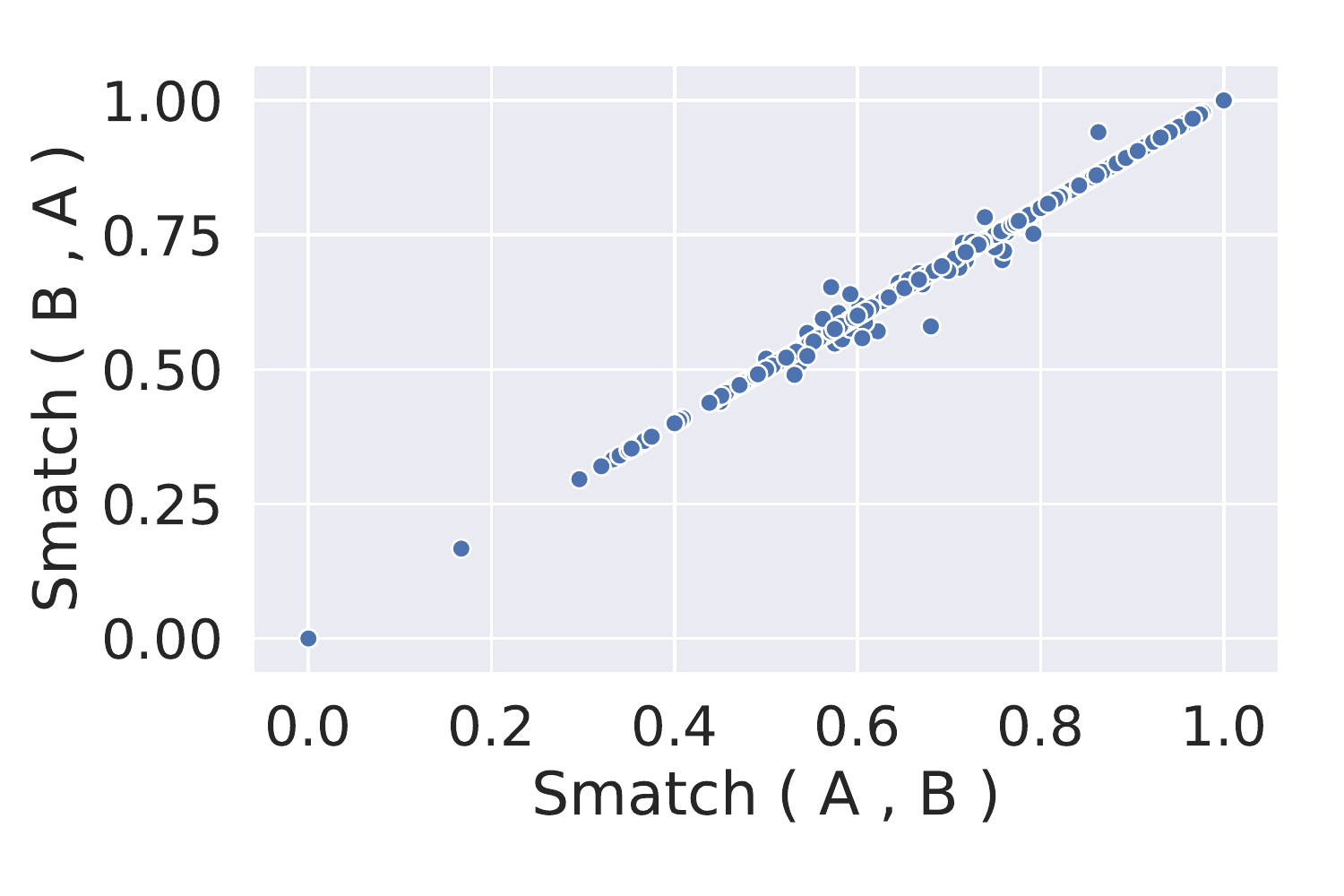}
  \caption{GPLA, Gold}
  \label{fig:sfig4}
\end{subfigure}
\begin{subfigure}{.33\textwidth}
  \centering
  \includegraphics[width=\linewidth,trim=0 0.85cm 0 0.95cm,clip]{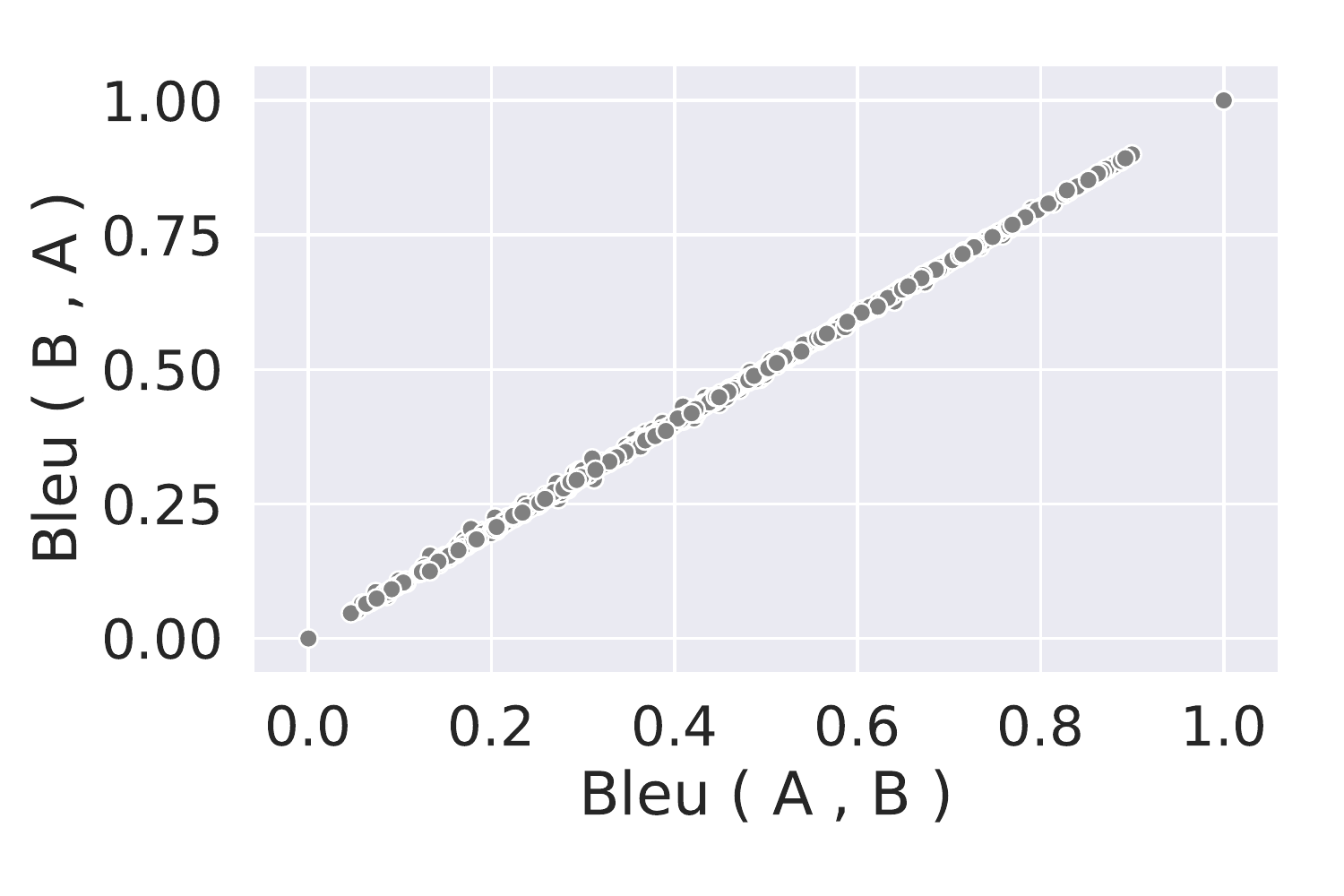}
  \caption{newstest2018, \textsc{Bleu}, avg. case}
  \label{fig:sfig3}
\end{subfigure}
\begin{subfigure}{.33\textwidth}
  \centering
  \includegraphics[width=\linewidth,trim=0 0.85cm 0 0.95cm,clip]{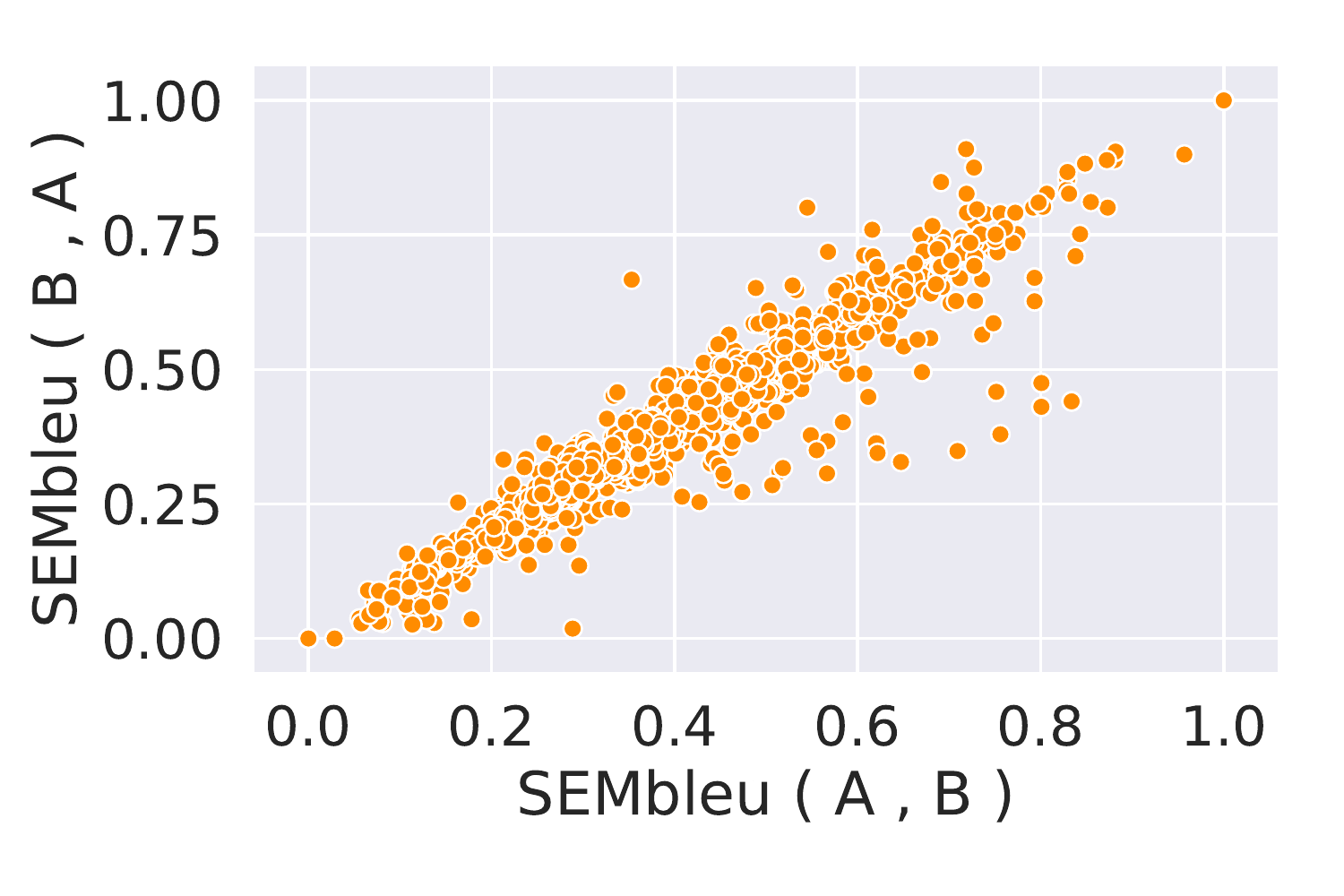}
  \caption{CAMR, Gold}
  \label{fig:sfig2}
\end{subfigure}
\begin{subfigure}{.33\textwidth}
  \centering
  \includegraphics[width=\linewidth,trim=0 0.85cm 0 0.95cm,clip]{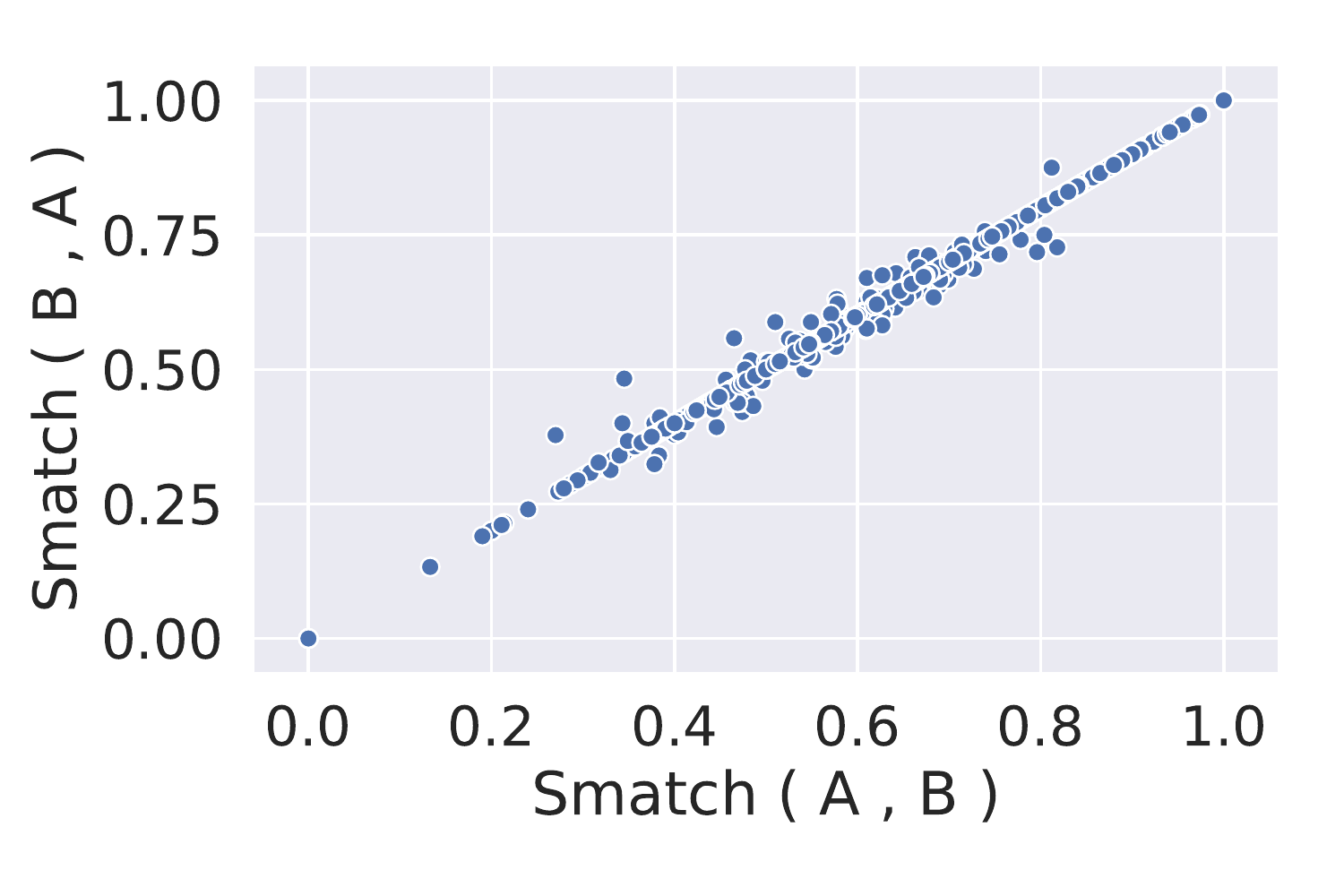}
  \caption{CAMR, Gold}
  \label{fig:sfig5}
\end{subfigure}
\begin{subfigure}{.33\textwidth}
  \centering
  \includegraphics[width=\linewidth,trim=0 0.85cm 0 0.95cm,clip]{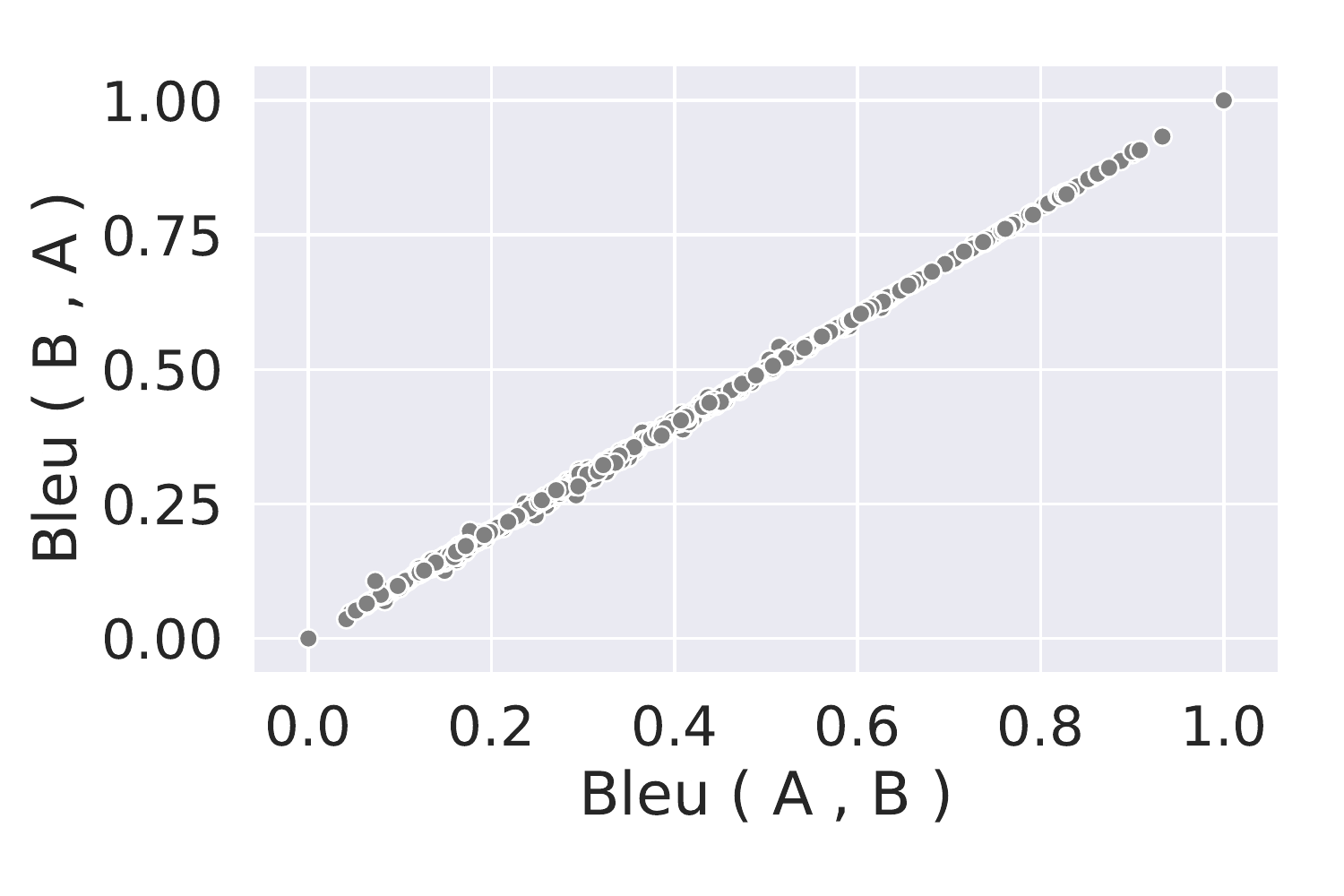}
  \caption{newstest2018, \textsc{Bleu}, worst case}
  \label{fig:sfig6}
\end{subfigure}
\caption{Symmetry evaluations of metrics. \textsc{SemBleu} (left column) and \textsc{Smatch} (middle column) and \textsc{Bleu} as a `baseline' in an MT task setting on newstest2018. \textsc{SemBleu}: large  divergence, strong outliers. \textsc{Smatch}: few divergences, few outliers; \textsc{Bleu}: many small divergences, zero outliers. 
(a) marks the case in Figure \ref{fig:violation-2}.
}
\label{fig:symm}
\vspace{-2mm}
\end{figure*}

This principle is fulfilled if $ \forall A,B \in \mathcal{D}: metric(A,B) = metric (B,A)$. Figure \ref{fig:violation-2} shows an example where 
\textsc{SemBleu} does not comply with this principle, to a significant extent: when comparing AMR graph $A$ against $B$, it yields a score greater than 0.8, yet, when comparing 
$B$ to 
$A$ the score is smaller than 0.5. 
We perform an experiment that quantifies this effect on a larger scale by assessing the frequency and the extent of such divergences.
To this end, we parse 1368 development sentences from an AMR corpus (LDC2017T10)
with an AMR parser (obtaining graph bank $\mathcal{A}$) and evaluate it against another graph bank $\mathcal{B}$ (gold graphs or another parser-output). We quantify the symmetry violation by the \textit{symmetry violation ratio} (Eq.\ \ref{eq:svr}) and the \textit{mean symmetry violation} (Eq. \ref{eq:msv}) given some metric $m$:

\vspace*{-4mm}
\begin{align}
    svr=\frac{\sum_{i=1}^{|\mathcal{A}|} \mathbb{I}[m(\mathcal{A}_i,\mathcal{B}_i) \ne m(\mathcal{B}_i,\mathcal{A}_i)]}{|\mathcal{A}|}
    \label{eq:svr}\\
    msv=\frac{\sum_{i=1}^{|\mathcal{A}|} |m(\mathcal{A}_i,\mathcal{B}_i) - m(\mathcal{B}_i,\mathcal{A}_i)|}{|\mathcal{A}|}
    \label{eq:msv}
\end{align}

We conduct the experiment with
three AMR systems, CAMR \cite{wang2016camr}, GPLA \cite{lyu-titov-2018-amr} and JAMR \cite{flanigan-etal-2014-discriminative}, and the gold graphs. 
Moreover, to provide a baseline that allows us to better put the results into perspective, we also estimate the symmetry violation of \textsc{Bleu} (\textsc{SemBleu}'s MT ancestor) in an MT setting. Specifically, 
we fetch 16 system outputs of the WMT 2018 EN-DE metrics task \cite{DBLP:conf/wmt/MaBG18} and calculate \textsc{Bleu}(A,B) and \textsc{Bleu}(B,A) of each sentence-pair (A,B) from the MT system's output and the reference  (using the same smoothing method as \textsc{SemBleu}). As \textit{worst-case}/\textit{avg.-case}, we use the outputs from the team where \textsc{Bleu} exhibits maximum/median $msv$.\footnote{worst: LMU uns.; avg.: LMU NMT \cite{huck-etal-2017-lmu}.}

\begin{table}[t]
    \centering
    \scalebox{0.68}{
    \begin{tabular}{@{}lrrrr}
    \toprule
     & \multicolumn{4}{c}{symmetry violation}\\
     \cmidrule{2-5}
    & \multicolumn{2}{c}{svr (\%, $\Delta$>0.0001)}& \multicolumn{2}{c}{msv (in points) }\\
    
    Graph banks & \textsc{Smatch}  & \textsc{SemBleu}  & \textsc{Smatch} &\textsc{SemBleu} \\
    \toprule
    Gold $\leftrightarrow$ GPLA & 2.7 & 81.8 &0.1 &3.2 \\
    Gold $\leftrightarrow$ CAMR & 7.8 & 92.8 & 0.2 & 3.1\\
    Gold $\leftrightarrow$ JAMR &5.0 & 87.0&0.1&3.2\\
    JAMR $\leftrightarrow$ GPLA & 4.2 & 86.0 & 0.1 &3.0 \\
    CAMR $\leftrightarrow$ GPLA & 7.4&93.4&0.1&3.4\\
    CAMR $\leftrightarrow$ JAMR & 7.9 &91.6 & 0.2& 3.3\\
    \midrule
    avg.\ &5.8 &88.8&0.1& 3.2\\

    \bottomrule
    \end{tabular}}
    \caption{svr (Eq.\ \ref{eq:svr}), msv (Eq.\ \ref{eq:msv}) of AMR metrics.\\ 
    }
    \label{tab:svrmsv}
\vspace{+2mm}
    \centering
    \scalebox{0.75}{
    \begin{tabular}{@{}lrr}
    \toprule
     & \multicolumn{2}{c}{\textsc{Bleu} symmetry violation, MT}\\
     \cmidrule{2-3}
   data: newstest2018 $\leftrightarrow (\cdot)$ & svr (\%, $\Delta$>0.0001)& msv (in points)\\
    
    \toprule
    worst-case & 81.3  & 0.2  \\
    avg-case & 72.7  & 0.2\\
    \bottomrule
    \end{tabular}}
    \caption{svr (Eq.\ \ref{eq:svr}), msv (Eq.\ \ref{eq:msv}) of \textsc{Bleu}, MT setting. 
    }
    \label{tab:msvbleu}
    \vspace{-2mm}
\end{table}
Table \ref{tab:svrmsv} shows that 
more than 80\% of the eva\-lu\-ated AMR graph pairs lead to a symmetry violation with \textsc{Sem\-Bleu} (as opposed to less than 10\% for \textsc{Smatch}). The $msv$ of \textsc{Smatch} is considerably smaller compared to \textsc{SemBleu}: 0.1 vs.\ 3.2  points F1 score. Even though the \textsc{Bleu} metric is inherently asymmetric, most of the 
symmetry violations are negligible when applied in MT (high $svr$, low $msv$, Table \ref{tab:msvbleu}). However, when applied to AMR graphs `via'
\textsc{SemBleu} the asymmetry is amplified by a factor of approximately 16 (0.2 vs.\ 3.2 points).
Figure  \ref{fig:symm} visualizes the sym\-me\-try violations of \textsc{SemBleu} (left), \textsc{Smatch} (middle) and \textsc{Bleu} (right). The \textsc{Sembleu}-plots show that the effect is widespread, some cases are extreme, many others are less extreme but still considerable. This stands in contrast to \textsc{Smatch} but also to \textsc{Bleu}, which itself appears well calibrated and does not suffer from any major asymmetry.

In sum, symmetry violations with \textsc{Smatch} are much fewer and less pronounced than those observed with \textsc{SemBleu}. In theory, \textsc{Smatch} is ful\-ly symmetric, however, vio\-la\-tions can occur due to alignment errors from the greedy variable-alignment search (we discuss this issue in the next paragraph). By contrast, the sym\-me\-try violation of \textsc{SemBleu} is intrinsic to the method since the underlying overlap measure \textsc{Bleu} is inherently asymmetric, however, 
this asymmetry is amplified in \textsc{SemBleu}  compared to \textsc{Bleu}.\footnote{As we show below (principle V), this is due to the 
way in which k-grams are extracted from variable-free AMR graphs.}
 
 \label{par:det}\paragraph{IV.\  Determinacy} This principle states that repeated calculations of a metric should yield identical results. Since there is no ran\-dom\-ness in \textsc{SemBleu}, it fully complies with this principle. The reference implementation of \textsc{Smatch} does not fully guarantee deterministic variable alignment results, 
 since it aligns the variables by means of greedy hill-climbing. However,
 multiple random initializations together
\begin{table}
    \centering
    \scalebox{0.73}{
    \begin{tabular}{lrrrrr}
    & \multicolumn{5}{c}{\# restarts}\\
    & 1 & 2 & 3  & 5 & 7\\
    \toprule
       corpus vs.\ corpus & 2.6e$^{-4}$ &  1.7e$^{-4}$ & 8.1e$^{-5}$& 5.7e$^{-5}$& 5.6e$^{-5}$  \\
       graph vs.\ graph  & 1.3e$^{-3}$ &  1.0e$^{-3}$ & 8.5e$^{-4}$& 5.3e$^{-4}$& 4.0e$^{-4}$\\
    \bottomrule
    \end{tabular}}
    \caption{Expected determinacy error $\epsilon$ in \textsc{Smatch} F1.}
    \label{tab:determinacye}
    \vspace{-3mm}
\end{table}
with the small set of AMR variables imply that the deviation will be $\leq \epsilon$ (a small number close to 0).\footnote{Additionally, $\epsilon=0$ is guaranteed when resorting to a (costly) ILP calculation \citep{cai-knight-2013-smatch}.} In Table \ref{tab:determinacye} we  measure the expected $\epsilon$: it displays the \textsc{Smatch} F1 standard deviation with respect to 10 independent runs, on a corpus level and on a graph-pair level (arithmetic mean).\footnote{Data: dev set of LDC2017T10, parses by GPLA.} We see that $\epsilon$ is small, even when only one random start is performed (corpus level: $\epsilon$=$0.0003$, graph level: $\epsilon$=$0.0013$). We conclude that the hill-climbing in \textsc{Smatch} is unlikely to have any significant effects on the final score.

\paragraph{V.\  No bias}
\label{sec:bias}
A  similarity metric of (A)MRs should
not unjustifiably or unintentionally favor the correctness or penalize errors pertaining to any (sub-)structures of the graphs. However, we find that \textsc{SemBleu} is affected by a bias that affects (some) leaf nodes attached to  high-degree nodes.
The bias arises from two related factors: (i) when transforming $G$ to $G^{vf}$, \textsc{SemBleu} replaces variable nodes with concept nodes.
Thus, nodes which were 
leaf nodes in $G$ can be rai\-sed  to highly connected nodes in $G^{vf}$. 
(ii)
breadth-first k-gram ex\-trac\-tion starts  from the root node. 
During graph traversal, 
concept leaves -- now occupying the position of (former) variable nodes 
with a high number of outgoing (and incoming) edges -- will be visited and extracted more frequently than others. 

\begin{figure}
    \centering
    \includegraphics[width=0.8\linewidth]{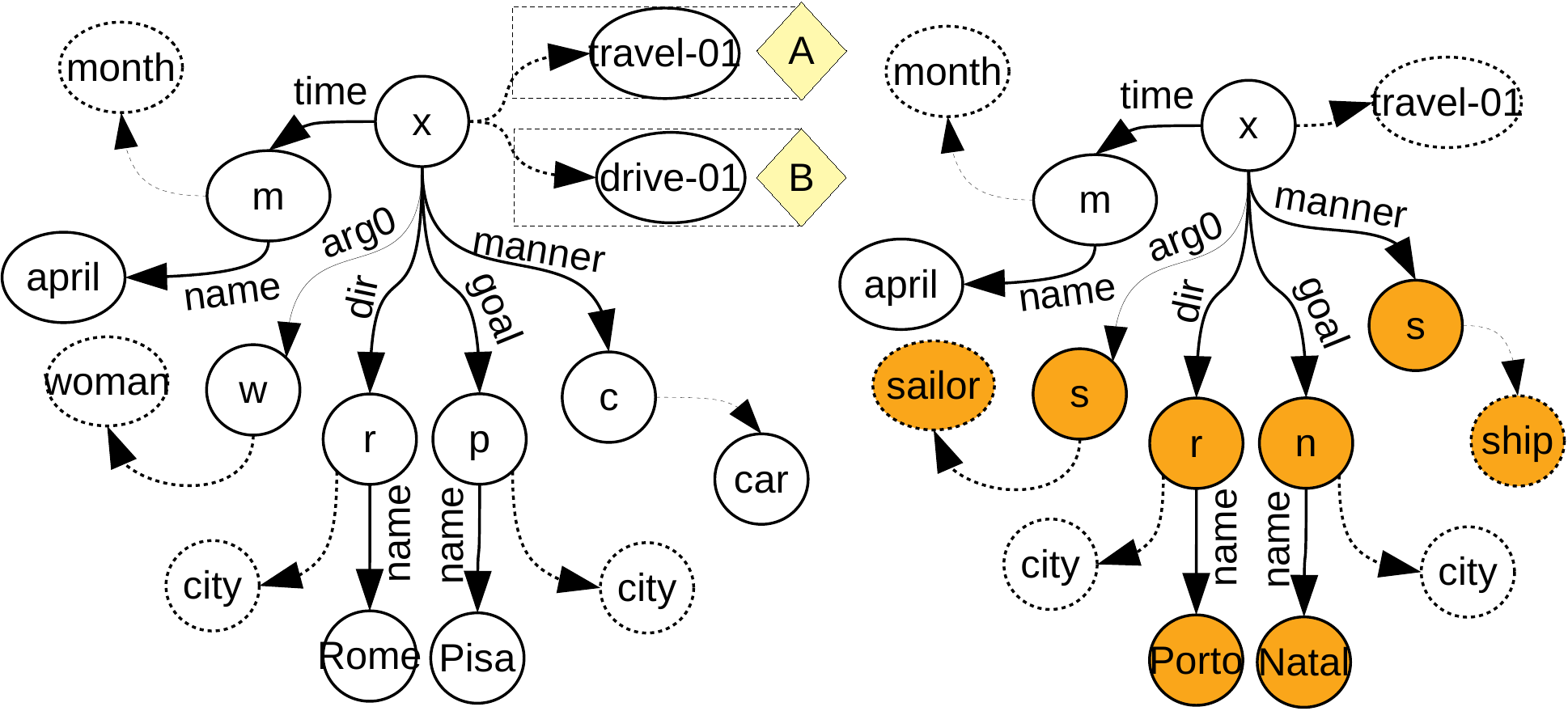} 
    \vspace{-3mm}
    \begin{ssmall}
\begin{Verbatim}[commandchars=\\\{\},codes={\catcode`$=3\catcode`^=7}]
\textbf{-----------------------Scores-----------------------}
\textbf{     metric (leftA,leftB)     metric (leftA,right)}
\textbf{       ---------------          ---------------} 
\textbf{       \textsc{SemBleu} -> 0.38     \textcolor{red}{<}    \textsc{SemBleu} -> 0.46}
\textbf{       \textsc{Smatch}  -> 0.87     \textcolor{cadmiumgreen}{>}    \textsc{Smatch}  -> 0.73}
\textbf{-----------------------------------------------------}
  \end{Verbatim}
\end{ssmall}
\vspace{-4mm}
    \caption{Left: \textit{In April, a woman rides a car from Rome to Pisa.} 
    root nodes
    A: \textit{travel-01} vs.\ B: \textit{drive-01}. Right: \textit{In Apr., a sailor travels with a ship from P. to N.} 
    }
    \label{fig:bias}
    \includegraphics[width=0.5\linewidth]{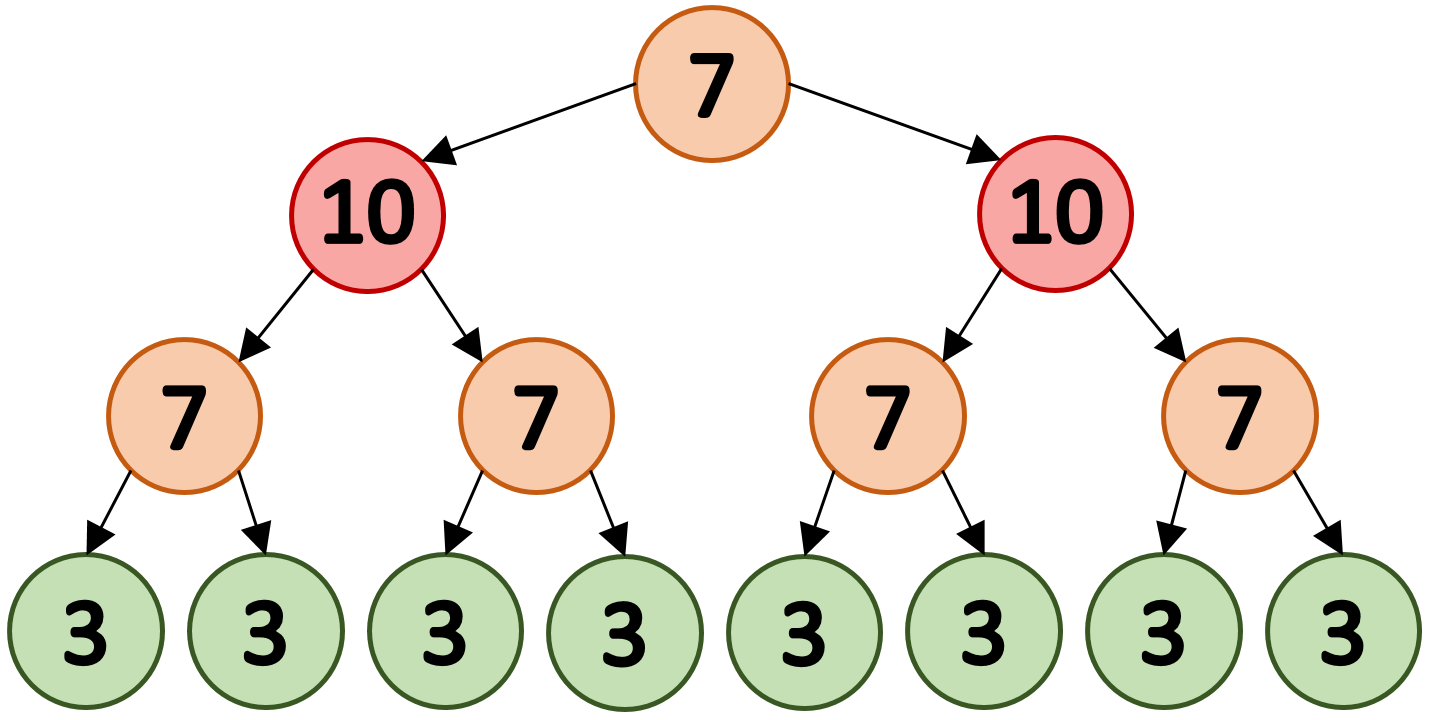}
    \caption{\# of $k$-grams entered by a node in \textsc{SemBleu}.}
    \label{fig:bias-heatmap}
\end{figure}

The two factors in com\-bi\-na\-tion make \textsc{SemBleu} penalize a wrong concept node harshly when it is attached to a high-degree variable node (the leaf is raised to high-\-degree when transforming $G$ to $G^{vf}$).
Conversely, correct or wrongly assigned concepts attached to nodes with low degree are only weakly considered.\footnote{This may have severe consequences, e.g., for \textit{negation}, since negation \textit{always} occurs as a leaf in $G$ and $G^{vf}$. 
Therefore, \textsc{SemBleu}, by-design, is benevolent to polarity errors.
} E.g., consider Figure \ref{fig:bias}. \textsc{SemBleu} considers two graphs that express quite distinct meanings (left and right) as more similar than graphs that are almost equivalent in meaning (left, variant A vs.\ B). This is because the leaf that is attached to the root is raised to a highly connected node in $G^{vf}$ and thus is over-frequently contained in the extracted k-grams, whereas the other leaves will remain leaves in $G^{vf}$.

\paragraph{Analyzing and quantifying \textsc{SemBleu}'s bias} To better understand the bias, we study three limiting cases: (i) the root is wrong 
(${}^{{}_\bsqrt{{}}}$)
 (ii) $d$ leaf nodes are wrong  ($\includegraphics[scale=1.05]{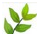}$) and (iii)  one branching node is wrong ($\includegraphics[scale=0.75]{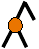}$). Depending on a specific node and its position in the graph, we would like to know onto how many k-grams (\textsc{SemBleu}) or triples (\textsc{Smatch}) the errors are projected. For the sake of simplicity, we assume that the graph always comes in its simplified form $G^{vf}$, that it is a tree, and that every non-leaf node has the same out-degree $d$. 

\begin{table}
\scalebox{0.95}{
\begin{tabular}{lrrr}
\toprule
      
      &   \includegraphics[scale=1.35]{leaf-crop.pdf} & ${~}^\bsqrt{~}$ &
      $\text{\includegraphics[scale=0.85]{pics/branch-twig-crop.pdf}}$\\
      \toprule
   $\textsc{SemBleu}$ & $\mathcal{O}(3d)$ & $\mathcal{O}(d^2+d)$ & $\mathcal{O}(d^2+2d)$ \\
   $\textsc{Smatch}$ & $\mathcal{O}(d)$ & $\mathcal{O}(d)$ & $\mathcal{O}(d)$ \\
   \bottomrule
   \end{tabular}}

   \caption{Error impact depending on error location in a tree with node degree $d$. 
   }
    \label{tab:scenarios}
    \vspace{-3mm}
\end{table}

The result of our analysis is given in Table \ref{tab:scenarios}\footnote{Proof sketch, \textsc{Smatch}, $d$ leaves: $d$ triples, a root: $d$ triples, a branching node:  $d$+$1$ triples. \textsc{SemBleu}$_{\text{k=3}}^{\text{w}_k\text{=}1/3}$, $d$ leaves: $3d$ k-grams ($d$ tri, $d$ bi, $d$ uni). A root:  $d^2$ tri, $d$ bi, 1 uni. A branching node: $d^2$+$d$+$1$ tri, $d$+$1$ bi, 1 uni.$\qed$} and exemplified in Figure \ref{fig:bias-heatmap}. Both show that the number of times k-gram extraction visits a node heavily depends on its position and that with growing $d$, the bias gets amplified (Table \ref{tab:scenarios}).\footnote{Consider that in AMR, $d$ 
 can be quite high, e.g., a predicate with multiple arguments and additional modifiers.} E.g., when $d$=3, 3 wrong leaves yield 9 wrong k-grams, and 1 wrong branching node can already yield 18 wrong k-grams. By contrast, in \textsc{Smatch} 
 the weight of $d$ leaves always approximates the weight of 1 branching node of  degree $d$. 
 
 In sum, in \textsc{Smatch} the impact of a wrong node is constant for all node types and rises linearly with $d$. In \textsc{SemBleu} the impact of a node rises approximately quadratically with $d$ and it also depends on the node type, since it raises some (but not all) leaves in $G$ to connected nodes in $G^{vf}$.

\paragraph{Eliminating biases} A possible approach to reduce \textsc{SemBleu}'s biases could be to weigh the extracted k-gram matches according to the degree of the contained nodes. However, this would imply that we assume some k-grams (and thus also some nodes and edges) to be of greater importance than others -- in other words, we would eliminate one bias by introducing another. Since the breadth-first traversal is the metric's backbone, this issue may be hard to address well. When \textsc{Bleu} is used for MT evaluation, there is no such bias because the k-grams in a sentence appear linearly.

\paragraph{VI.\  Graph matching: symbolic perspective}\label{subsec:alignornot}
This principle requires that a metric's score grows with increasing overlap of the conditions that are simultaneously contained in $A$ and $B$. \textsc{Smatch} fulfills this principle since it matches two AMR graphs inexactly \cite{yan2016short,riesen2010exact} by aligning variables s.t.\ that the triple matches are maximized. 
Hence, \textsc{Smatch} can be seen as a graph mat\-ch\-ing algorithm that works on any pair of graphs that contain (some) nodes that are variables. It fulfills the Jaccard-based overlap objective which symmetrically measures the amount of triples on which two graphs agree, normalized by their respective sizes  (since \textsc{Smatch} F1 = $2J/(1+J)$ is a monotonic relation).

Since \textsc{SemBleu} does not satisfy principles II and III (id.\ of indescernibles and symmetry), it is a co\-rol\-lary that it cannot fulfill the overlap objective.\footnote{Proof by symmetry violation:\\ w.l.o.g.\ $\exists A,B$: $metric(A,B)$ $>$ $metric(B,A)$ $\Rightarrow$ $f(A,B)$ $>$ $f(B,A)$ $\rightarrow$ ~\lightning~, since $f(A,B) = |t(A) \cap t(B)| = |t(B) \cap t(A)| = f(B,A)\qed$ /// Proof by identity of indiscernibles:\\ w.l.o.g.\ $\exists$ $A,B,C:$ $metric(A,B)$ $=$ $metric(A,C)$ $=$ $1$ $\land$ $f(A,B)/z(A,B)$ = 1 $>$  $f(A,C)/z(A,C)$\lightning$\qed$
} 
  Generally, \textsc{SemBleu} does not com\-pare and match two AMR graphs per se, instead it matches the results of a graph-to-bag-of-paths projection function (\S \ref{par:sembleudecript}) and the input may not be recoverable from the output (surjective-only). Thus, mat\-ch\-ing the outputs of this function cannot be equated to matching the inputs on a graph-level.

\section{Towards a more \textit{semantic} metric for \textit{semantic} graphs: \textsc{S$^2$match}}
\label{sec:s2}
\begin{figure}
    \centering
    \includegraphics[width=\linewidth,trim=0 0.007cm 0 0.03cm,clip]{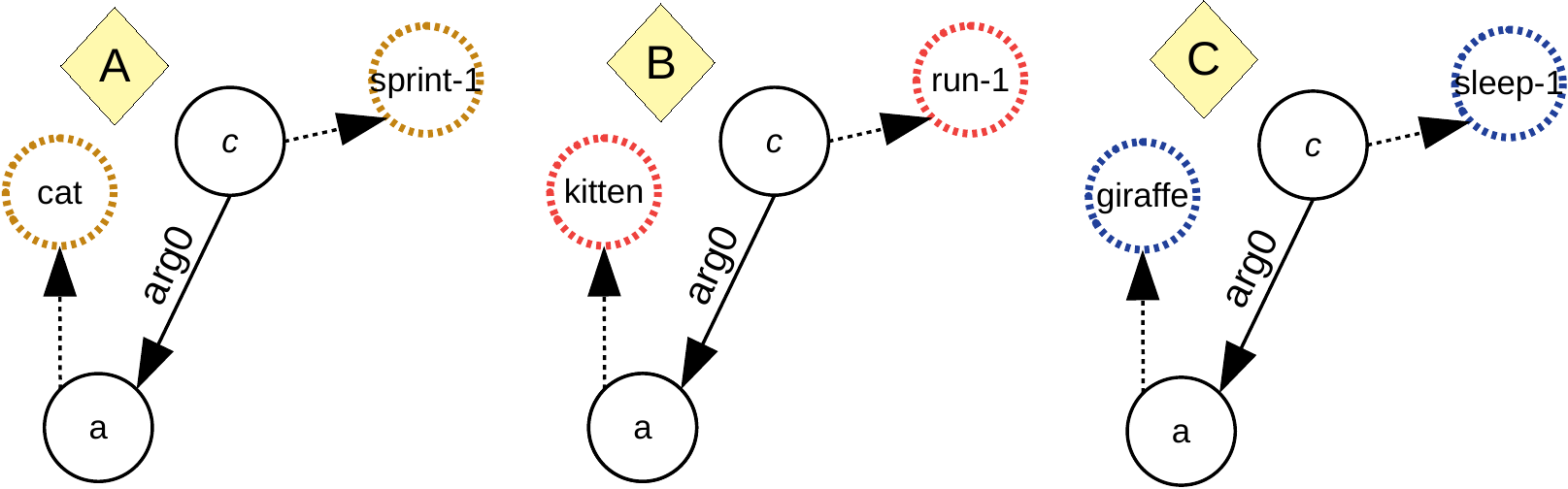}
    \begin{ssmall}
    \vspace{-2mm}
\begin{Verbatim}[commandchars=\\\{\},codes={\catcode`$=3\catcode`^=7}]
\textbf{-----------------------Scores-----------------------}
\textbf{  metric (A,B)       metric (B,C)       metric (A,C)}
\textbf{---------------    ---------------    ---------------}
\textbf{\textsc{SemBleu} -> 0.00    \textsc{SemBleu} -> 0.00    \textsc{SemBleu} -> 0.00}
\textbf{\textsc{Smatch}  -> 0.25    \textsc{Smatch}  -> 0.25    \textsc{Smatch}  -> 0.25}
\textbf{\textsc{S$^2$match} -> 0.39    \textsc{S$^2$match} -> 0.25    \textsc{S$^2$match} -> 0.25 }
\textbf{-----------------------------------------------------}
   \end{Verbatim}
   \vspace{-2mm}
\end{ssmall}
    \caption{Three different AMR graphs representing \textit{The cat sprints; The kitten runs; The giraffe sleeps} and pairwise similarity scores from \textsc{SemBleu, Smatch} and \textsc{S$^{2}$Match} (see (\S \ref{sec:s2}) for S$^2$Match).
    } 
    \label{fig:catkitten}
    \vspace{-2mm}
\end{figure}

This section focuses on  principle VII, semantically graded graph matching, a principle that
none of the 
AMR metrics considered so-far 
satisfies. A fulfilment of this principle also 
increases the capacity of a metric to assess the semantic similarity of two AMR graphs from \textit{different sentences}. E.g., when clustering AMR graphs or detecting paraphrases in AMR-parsed texts, the ability to ab\-stra\-ct away from concrete lexicalizations is clearly de\-si\-ra\-ble. Consider Figure \ref{fig:catkitten} with three different graphs. Two of them $(A,B)$ are similar in meaning and differ significantly from $C$. However, both \textsc{Smatch} and \textsc{SemBleu} yield the same result in the sense that $metric(A,B) = metric(A,C)$. Put differently, neither metric takes into account that \textit{giraffe} and \textit{kitten} are 
two quite different concepts, while \textit{cat} and \textit{kitten} are more similar. 
However, we would like this to be reflected by our metric and obtain $metric(A,B)>metric(A,C)$ in such a case. 

\paragraph{\textsc{S$^2$match}} We propose the \textsc{S$^2$match} metric (\textit{\underline{S}oft \underline{S}emantic \underline{match}}, pronounced:
\textipa{[estu:m\ae t\textesh]}) that builds on 
\textsc{Smatch} 
but differs from it
in one important aspect: instead of maximizing the number of (hard) triple matches between two graphs during alignment search, we maximize the (soft) triple matches by taking into account the semantic similarity of concepts. Recall that an AMR graph contains two 
types of triples: instance
and relation triples (e.g., Figure \ref{fig:catkitten}, left: 
\itriple{a}{instance}{cat} and \rtriple{c}{arg0}{a}). In \textsc{Smatch}, two triples can only be matched 
if they are identical. In \textsc{S$^2$match}, we relax this constraint, which has also the potential to yield a different, and possibly, a better variable alignment. 
More precisely, in \textsc{Smatch} we match two instance triples \itriple{a}{instance}{x} $\in A$ and \itriple{map(a)}{instance}{y} $\in B$ as follows:
\begin{equation}
 hardMatch =  \mathbb{I}[\text{x} = \text{y}]
\end{equation}
where $\mathbb{I}(c)$ equals 1 if $c$ is true and $0$ otherwise.
\textsc{S$^{2}$match} relaxes this condition:
\begin{equation}
 softMatch =  1-d(\text{x}, \text{y}),
 \label{eq:sim}
\end{equation}

where $d$ is an arbitrary distance function $d: X \times X \rightarrow [0,1]$. E.g., in practice, if we represent the concepts as vectors $x,y \in \mathbb{R}^n$, we can use 
\begin{equation}
    d(x,y) = \min \bigg\{ 1, 1 - \frac{y^T x}{\norm{x}_2 \norm{y}_2} \bigg\}. 
    \label{eq:cos}
\end{equation} 

When plugged into Eq.\ \ref{eq:sim}, this results in the \textit{cosine similarity} $\in [0,1]$. It may be suitable to set a threshold $\tau$ (e.g., $\tau = 0.5$), to only consider the similarity between two concepts if it is above $\tau$ ($softMatch=0$ if $1-d(x,y) < \tau$). In the following pilot experiments, we use cosine (Eq.\ \ref{eq:cos}) and $\tau = 0.5$ over 100 dimensional GloVe vectors \cite{pennington2014glove}.

To summarize, \textsc{S$^2$match} is designed to either yield the same score as  \textsc{Smatch} --
 or a slightly increased score when it aligns concepts that are symbolically distinct but semantically 
similar. 
\begin{figure}[t]
    \centering
    \includegraphics[width=0.65\linewidth]{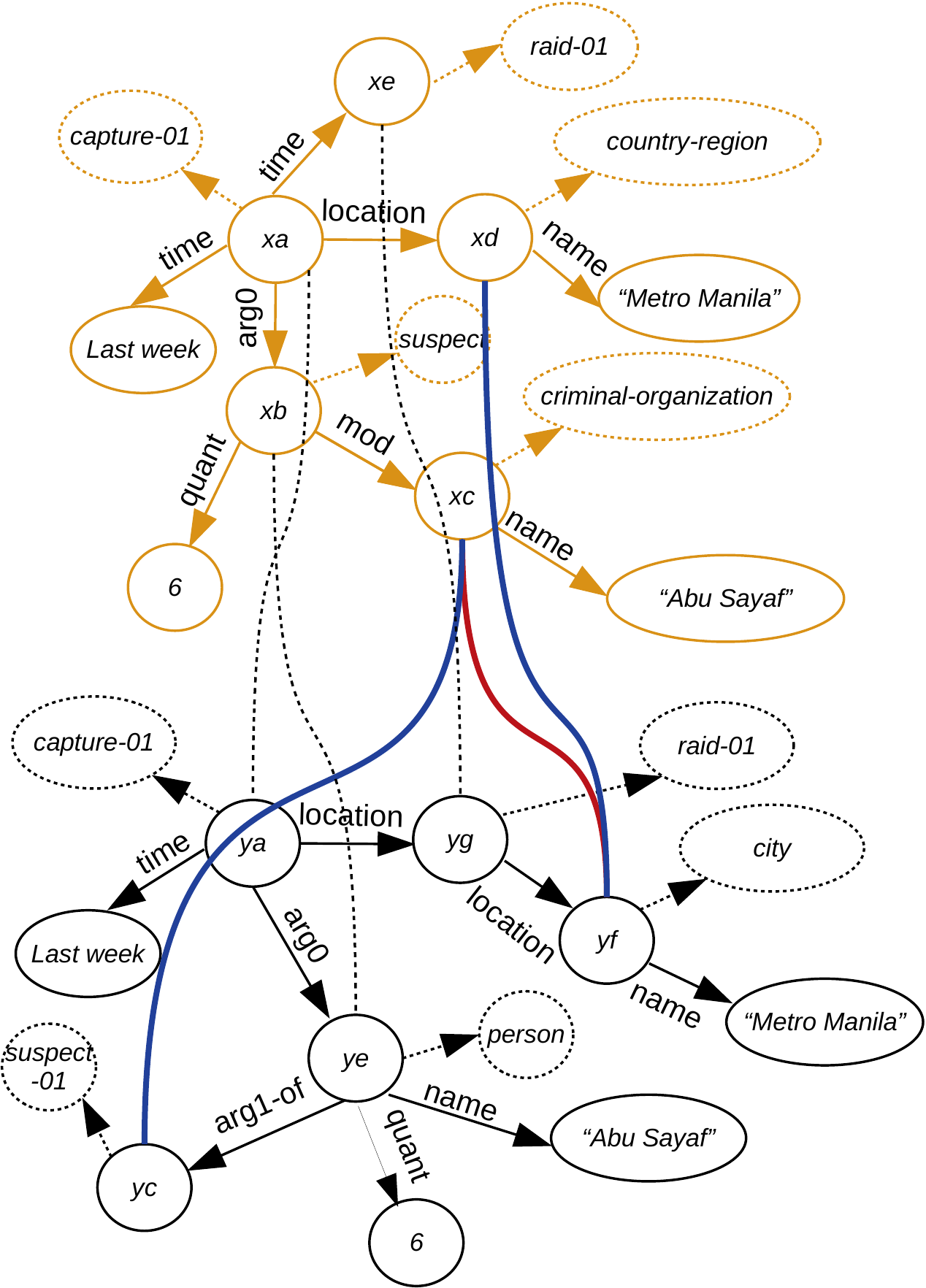}
    \caption{\textit{`6 Abu Sayyaf suspects were captured last week in a raid in Metro Manila.'} \textcolor{darkgoldenrod}{gold} (top) vs.\ parsed AMR (bottom). \textsc{Smatch} aligns \textit{criminal-organization} to  \textit{city} (\textcolor{red}{red}); \textsc{S$^2$match} aligns \textit{criminal-organization} to  \textit{suspect-01}, \textit{city} to \textit{country-region} (\textcolor{blue}{blue}).}
    \label{fig:newalign}
    \vspace{-3mm}
\end{figure}
\begin{table}
    \centering
    \scalebox{0.7}{
    \begin{tabular}{l|r|rrr}
    & & \multicolumn{3}{c}{determinacy error}\\
          & avg.\ msv (Eq.\ \ref{eq:msv}) & 1 restart & 2 restarts & 4 restarts\\
         \midrule
         \textsc{Smatch} & 0.0011 & 1.3e$^{-3}$ &1.0e$^{-3}$ &  5.3e$^{-4}$\\
         \textsc{S$^2$match}& 0.0005 & 9.0e$^{-4}$ & 6.1e$^{-4}$ & 2.1e$^{-4}$\\
         \midrule
         relative change & -54.6\% & -30.7\% & -39.0\% &-60.3\%\\
         \bottomrule
    \end{tabular}}
    \caption{\textsc{S$^2$match} improves upon \textsc{Smatch} by reducing the extent of its non-determinacy.}
    \label{tab:errorreduct}
    \vspace{-3mm}
\end{table}
\begin{table*}
    \centering
    \scalebox{0.58}{
    \begin{tabular}{@{}llllrl@{}}
    \toprule
        input span region (excerpt) &  amr region gold (excerpt) & amr region parser (excerpt) & cos&  points F1$\uparrow$& annotation \\
        \midrule
        40 km southwest of  & :quant 40 :unit (\colorbox{yellow}{\textbf{\textcolor{blue}{k2}} / kilometer}) & (\colorbox{yellow}{\textbf{\textcolor{blue}{k22}} / km} :unit-of (d23 / distance-quantity & 0.72& 1.2 &ex.\ similar\\
        improving agricultural prod.\ & (i2 / improve-01 ... :mod (\colorbox{yellow}{\textbf{\textcolor{blue}{f2}} / farming}) & (i31 / improve-01  :mod (\colorbox{yellow}{\textbf{\textcolor{blue}{a23}} / agriculture}) & 0.73 & 3.0 &ex.\ similar \\
        other deadly bacteria & op3 (\colorbox{yellow}{\textbf{\textcolor{blue}{b}} / bacterium} ... :mod (o / other))) & op3 (\colorbox{yellow}{\textbf{\textcolor{blue}{b13}} / bacteria} :ARG0-of  :mod (o12 / other)))& 0.80 &5.1 &ex.\ similar\\
        drug and law enforcement aid & (a / and :op2 (\colorbox{yellow}{\textbf{\textcolor{blue}{a3}} / aid-01} &:ARG1 (a9 / and :op1 (\colorbox{yellow}{\textbf{\textcolor{blue}{d8}} / drug}) :op2 (l10 / law))) & 0.67 & 1.8 & similar\\
        Get a lawyer and get a divorce. & :op1 (g / get-01 :mode imp.\ :ARG0 (\colorbox{yellow}{\textbf{\textcolor{blue}{y}} / you}) & :op1 (\colorbox{yellow}{\textbf{\textcolor{blue}{g0}} / get-01} :ARG1 (l2 / lawyer) :mode imp.) & 0.80 & 4.8 & dissimilar \\
        The unusual development. & ARG0 (d / develop-01  :mod (\colorbox{yellow}{\textbf{\textcolor{blue}{u}} / usual} :polarity -)) & :ARG0 (d1 / develop-02 :mod (\colorbox{yellow}{\textbf{\textcolor{blue}{u0}} / unusual}))& 0.60 & 14.0 & dissimilar\\
        \bottomrule
    \end{tabular}}
    \caption{Examples where \textsc{S$^2$match} assigns a higher score, 
    accounting for the similarity of \colorbox{yellow}{aligned concepts}. 
    }
    \label{tab:examples}
    \vspace{-3mm}
\end{table*}
An example, from parser evaluation, is shown in Figure \ref{fig:newalign}. 
Here,
\textsc{S$^2$match} increases the score to 63 F1 (+10 points) by detecting a more adequate alignment that accounts for the graded similarity of 
two related AMR 
concepts pairs. We believe that this is 
justified: The two graphs are very similar and an F1 of 53 is too low, doing the parser injustice.

On a technical note, the changes in alignments also have the outcome that \textsc{S$^2$match} mends some of \textsc{Smatch}'s flaws: It better addresses principles III and IV, reducing the symmetry violation and determinacy error (Table \ref{tab:errorreduct}).

\paragraph{Qualitative study: Probing \textsc{S$^2$match}'s choices} This study
randomly samples 100 graph pairs 
from 
those where \textsc{S$^2$match} assigned  
higher scores than \textsc{Smatch}.\footnote{Automatic graphs 
by 
GPLA, on LDC2017T10, dev set.} 
Two annotators were asked
to judge 
the similarity of all
aligned concepts with similarity score $<$1.0:
Are the concepts dissimilar, similar or extremely similar? 
For concepts 
judged dissimilar, we conclude that \textsc{S$^2$match} erroneously increased the score; if jud\-ged as
(extremely) similar, we conclude that 
the decision
was justified. We calculate three agreement statistics that all show large consensus among our annotators (Cohen's kappa equals 0.79, squared kappa: 0.87, Pearson's $\rho$:  0.91)
According to the annotations, 
the decision to increase the score is mostly justified: in 56\% and 12\% of cases both annotators voted that the newly aligned concepts are \textit{extremely similar} and \textit{similar}, respectively, while the agreed \textit{dissimilar} label makes up 25\% of cases. 

Table \ref{tab:examples} lists examples of good or ill-founded score increases. We observe, e.g., that \textsc{S$^2$match} accounts for the similarity of two concepts of different number: \textit{bacterium} (gold) vs.\ \textit{bacteria} (parser) (line 3). It also captures abbreviations (\textit{km -- kilometer}) and closely related concepts (\textit{farming -- agriculture}).
\textsc{SemBleu} and \textsc{Smatch} would
penalize the corresponding triples  in exactly the same way as predicting a truly dissimilar concept.

An interesting case is seen in line 7.
Here, \textit{usual} and \textit{unusual} are correctly annotated as dissimilar, since they are opposite concepts. 
\textsc{S$^2$match}, equipped with GloVe embeddings, measures a cosine of 0.6,
above the chosen threshold,
which results in an increase of the score by 14 points (the increase is large as these two graphs are tiny). It is well-known that synonyms and antonyms are difficult to dis\-tin\-guish with  distributional word representations, since they often share similar contexts. However, the case at hand is orthogonal to this problem: \textit{usual} in the gold graph is modified with the polarity `$-$', whereas the predicted graph assigned the (non-\-negated) opposite concept \textit{unusual}. Hence, given
the context in the gold graph, the prediction is semantically almost equivalent. This points to an aspect of principle VII 
that is not yet covered by \textsc{S$^2$match}: it assesses graded similarity at the lexical, but not at the phrasal level, and hence cannot account 
for compositional phenomena. 
In future work, we aim at 
alleviating 
this issue by 
developing extensions that measure semantic similarity for larger graph contexts, in order to fully satisfy all seven principles.\footnote{As we have seen, this requires much care. We therefore  consider this next step to be out of scope of the present paper.}

\paragraph{Quantitative study: metrics vs.\ human raters} 
This study investigates to what extent the judgments of the three metrics under discussion resemble human judgements, based on the following \textbf{two expectations}. First, the more a human rates two sen\-ten\-ces to be semantically \textit{similar} in their \textit{meaning}, the higher the metric should rate the cor\-res\-pon\-ding AMR graphs (\textbf{meaning similarity}). Second, the more a human rates two sen\-ten\-ces to be \textit{related} in their \textit{meaning} (maximum: equi\-va\-lence), the higher the score of our metric of the cor\-res\-pon\-ding AMR graphs should tend to be (\textbf{meaning relatedness}). Albeit not the exact same \cite{budanitsky-hirst-2006-evaluating}, the tasks are closely related and both in conjunction should allow us to better assess the performance of our AMR metrics.

As ground truth for the \textbf{meaning similarity} rating task we use 
test
data of the  Semantic Textual Similarity \textbf{(STS)}  shared task \cite{cer-etal-2017-semeval}, 
with 1,379 sentence pairs annotated for meaning similarity. For the \textbf{meaning-relatedness} task we use \textbf{SICK} \cite{marelli-etal-2014-semeval} with 9,840 sentence pairs that have been additionally annotated for semantic relatedness.\footnote{An example from SICK. Max.\ score: \textit{A man is cooking pancakes}--\textit{The man is cooking pancakes.}. Min.\ score: \textit{Two girls are playing outdoors near a woman.}--\textit{The elephant is being ridden by the man.} To further enhance the soundness of the SICK experiment we discard pairs with a \textit{contradiction} relation and retain 8,416 pairs with \textit{neutral} or \textit{entailment}.} We proceed as follows: We normalize the human ratings to [0,1]. Then we apply GPLA to parse the sentence tuples $(s_i,s'_i)$, obtaining tuples $(parse(s_i),parse(s'_i))$ and 
score the graph pairs with
the 
metrics:
\textsc{Smatch}(i), \textsc{S$^{2}$match}(i), \textsc{SemBleu}(i) and H(i),  where H(i) is the human score. 
\begin{table}[]
    \centering
    \scalebox{0.53}{
    \begin{tabular}{@{}l|lll|lll|lll|lll@{}}
    &\multicolumn{3}{c}{RMSE}  & \multicolumn{3}{c}{RMSE (quant)} & \multicolumn{3}{c}{Pearson's $\rho$}& \multicolumn{3}{c}{Spearman's $\rho$} \\
    \cmidrule{2-13}
       task & \textsc{SB} & \textsc{SM}& \textsc{S$^2$M} & \textsc{SB} & \textsc{SM}& \textsc{S$^2$M} & \textsc{SB} & \textsc{SM}& \textsc{S$^2$M} & \textsc{SB} & \textsc{SM}& \textsc{S$^2$M}\\
        \cmidrule{1-13}
        STS & 0.34& 0.25 &0.25 &0.25 & 0.11 & 0.10&0.52&0.55&0.55 & 0.51 & 0.53 & 0.53\\
        SICK & 0.38& 0.25 &0.24 & 0.32 &0.14 & 0.13 & 0.62 &0.64&0.64  &0.66 & 0.66 & 0.66 \\
        \bottomrule
    \end{tabular}}
    \caption{RMSE (lower is better) and correlation results of our metrics in our \textbf{STS} and \textbf{SICK} investigations. RMSE (quant): RMSE on empirical quantile distribution with quantiles 0.1,0.2,...,0.9.}
    \label{tab:stsvssick}
    \vspace{-2mm}
\end{table}
For both tasks 
\textsc{Smatch} and \textsc{S$^2$match} yield better or equal correlations with human raters than \textsc{SemBleu} (Table \ref{tab:stsvssick}). When considering the RMS error $\sqrt{n^{-1}\sum_{i=1}^n (H(i) - metric(i))^2}$.
the difference is even more pronounced.
\begin{figure}
    \centering
    \includegraphics[width=.52\linewidth,trim=0 0.25cm 0 0.25cm,clip]{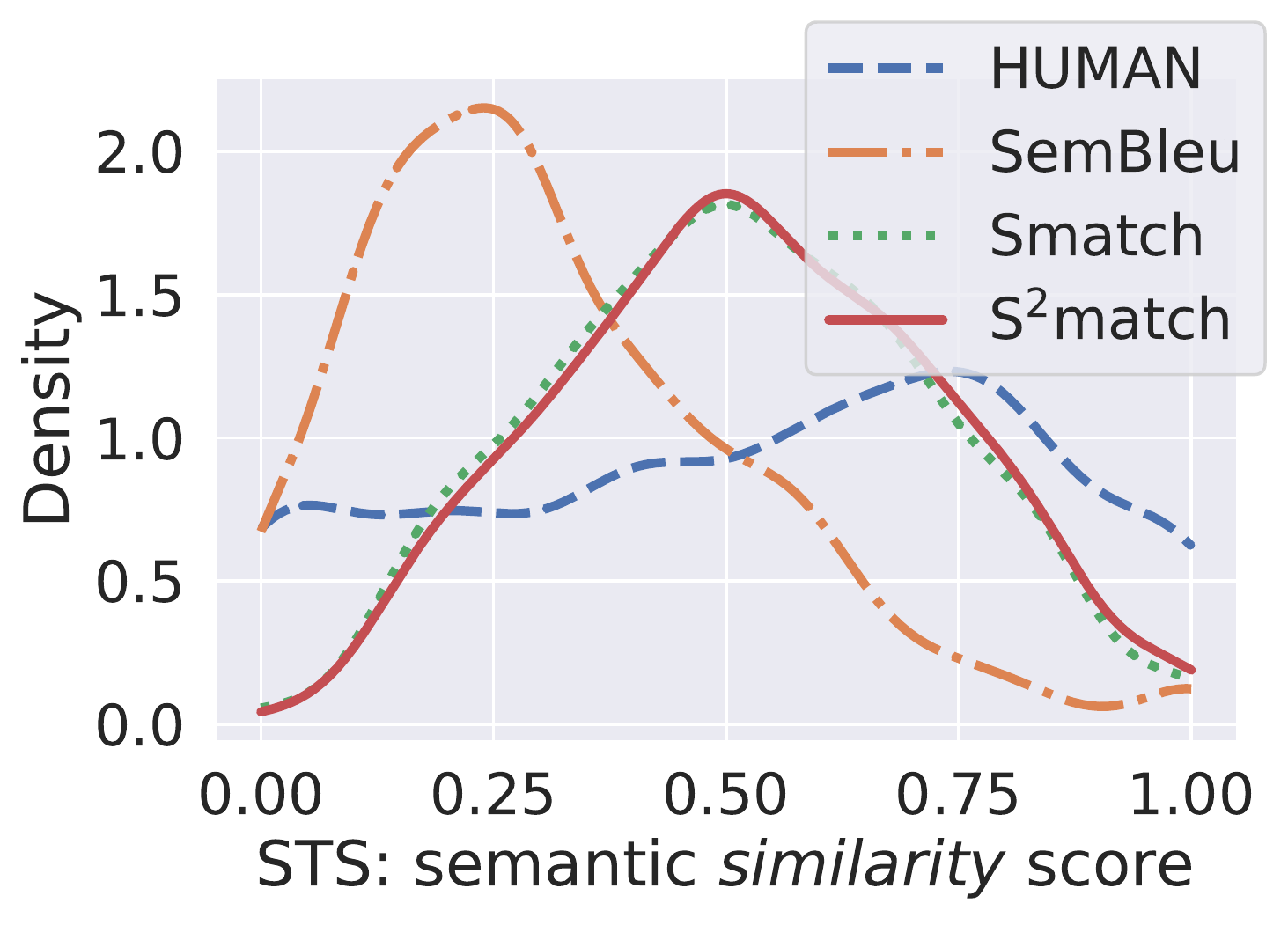}\includegraphics[width=.5\linewidth,trim=0 0.25cm 0 0.25cm,clip]{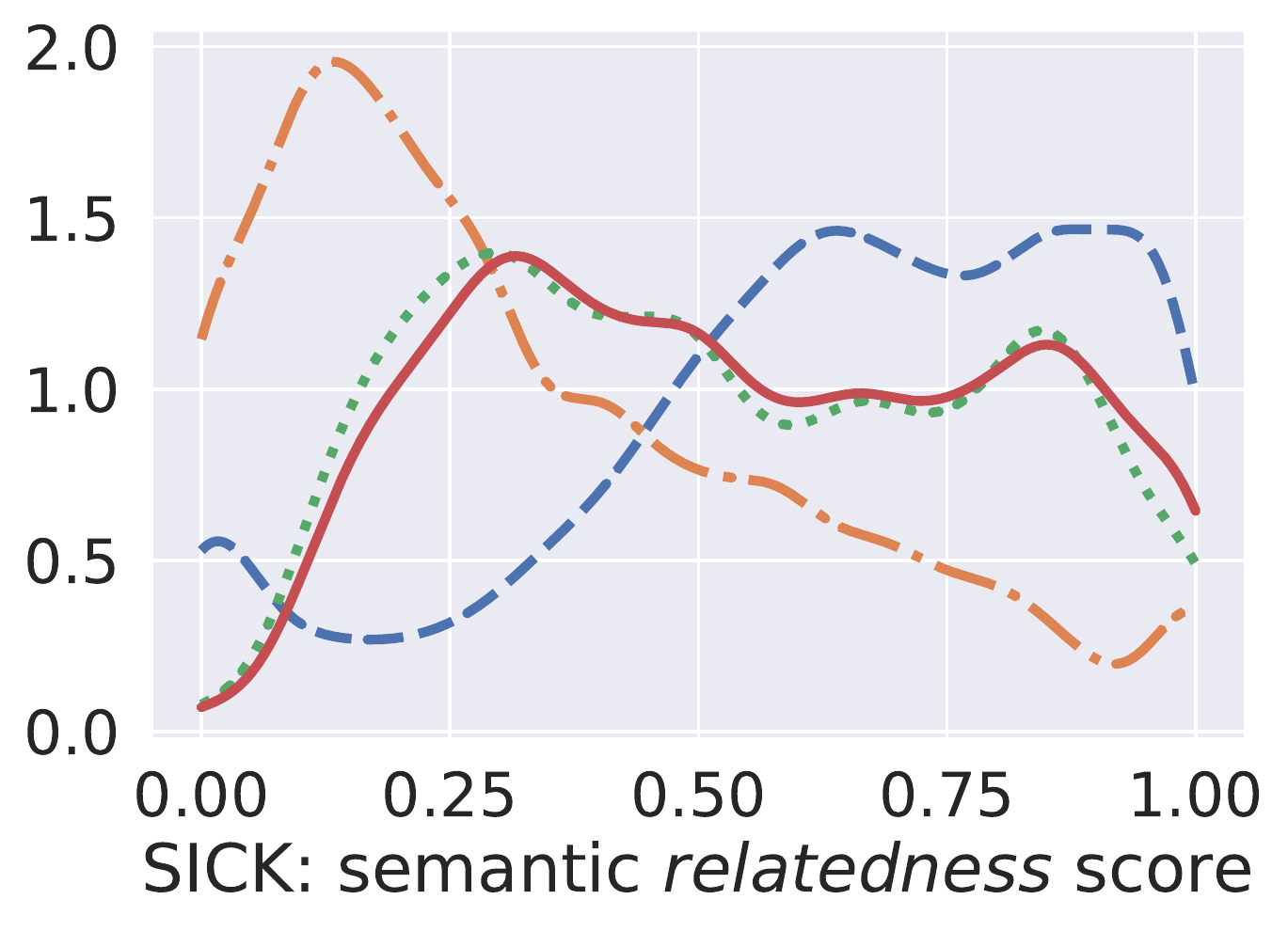}
    \caption{Sentence meaning similarity distributions.
    }
    \label{fig:sdist}
    \vspace{-2mm}
\end{figure}
\begin{figure}

\begin{ssmall}
\begin{Verbatim}[commandchars=\\\{\},codes={\catcode`$=3\catcode`^=7}]
\textbf{------------------Input Sentences-------------------}

\textbf{\textit{This is not a good idea.} | \textit{But it is not a good idea.}}

\textbf{------------\colorbox{lemonchiffon}{Human Similarity Judgement}-------------}

\textbf{\colorbox{lemonchiffon}{0.8} ("very similar, but not equivalent")}

\textbf{-----------------------Parses------------------------}

\textbf{(g2 / good-02           (c0 / contrast-01}
\textbf{  :ARG1 (i3 / idea        :ARG2 (i4 / idea}
\textbf{    \colorbox{lavender(web)}{:domain (t0 / this))}    \colorbox{lavender(web)}{:domain (it / it)}}
\textbf{  :polarity -)              :ARG1-of (g3 / good-02}
\textbf{                            :polarity -)))}

\textbf{------------\colorbox{lemonchiffon}{Metric Similarity Judgement}-----------}

\textbf{\textsc{SemBleu} = \colorbox{lemonchiffon}{0.38} < \textsc{Smatch} = \colorbox{lemonchiffon}{0.63} < \textsc{S$^2$match} = \colorbox{lemonchiffon}{0.74}}

\textbf{-----------------------------------------------------}
   \end{Verbatim}
   \end{ssmall}
\caption{An example from STS, where \textsc{S$^2$match} yields a score that better reflects the human judgement, due to detecting a similarity between the abstract anaphora \colorbox{lavender(web)}{\textit{it}} and \colorbox{lavender(web)}{\textit{this}}.}
\label{fig:updatedsimilarity}
\vspace{-3mm}
\end{figure}
This deviation in the absolute scores is also reflected by
the score density distributions plotted in Figure \ref{fig:sdist}: 
\textsc{Sembleu} underrates a good proportion of 
graph pairs 
whose
input sentences were rated as highly semantically similar or related by humans.
This may well relate to the biases of different node types (cf.\ \S \ref{sec:bias}). Overall \textsc{S$^2$match} appears to  provide a better fit with the score-\-distribution of the human rater when mea\-su\-ring \textbf{semantic similarity} and 
\textbf{relatedness}, 
the latter being 
notably closer to the human reference in some regions than the otherwise similar \textsc{Smatch}. A concrete example from the STS data is given in Figure \ref{fig:updatedsimilarity}. Here, \textsc{S$^2$match}\ detects the similarity between the abstract anaphors \textit{it} and \textit{this} and 
assigns a score that better reflects the human score compared to \textsc{Smatch} and  \textsc{SemBleu}, the latter being far too low. However, in total, we conclude that \textsc{S$^2$match}'s improvements seem rather small and no metric is perfectly aligned with human scores, possibly because gradedness of 
semantic similarity that arises in combination with constructional variation is not yet 
captured -- more
research is 
needed to extend \textsc{S$^2$match}'s scope to such cases.

\section{Metrics' effects on parser evaluation}
We have seen that different metrics can assign different scores to the same pair of graphs. We now want to assess to what extent this affects rankings: Does one metric rank a graph higher or lower than the other? And can this affect the ranking of parsers on benchmark datasets?

\paragraph{Quantitative study: graph rankings} 
In this 
experiment, we assess whether our metrics rank graphs differently. We use LDC2017T10 (dev) parses by CAMR $[c_1...c_n]$, JAMR $[j_1...j_n]$ and gold graphs $[y_1...y_n]$. Given  metrics $\mathcal{F}$ and $\mathcal{G}$ we obtain results $\mathcal{F}^{C}:=[\mathcal{F}(c_1,y_1) ... \mathcal{F}(c_n,y_n)]$ and analogously $\mathcal{F}^{J}$, $\mathcal{G}^{C}$ and $\mathcal{G}^{J}$. We calculate two statistics: (i) the ratio of cases $i$ where the metrics differ in their preference for one parse over the
other $(\mathcal{F}^{J}_i - \mathcal{F}^{C}_i)\cdot (\mathcal{G}^{J}_i - \mathcal{G}^{C}_i) < 0$, and, to assess significance, (ii) a t-test for paired samples on the differences assigned by the metrics
between the parsers: $\mathcal{F}^{J} - \mathcal{F}^{C}$ and $\mathcal{G}^{J} - \mathcal{G}^{C}$.

\begin{table}
\centering
\scalebox{0.71}{
\begin{tabular}{@{}l@{~}rr|rr|rr@{}}
\toprule
     & \textsc{Sm}$_{A,B}^{G}$& \textsc{Sm}$_{G}^{A,B}$ & \textsc{SB}$_{A,B}^{G}$ & \textsc{SB}$_{G}^{A,B}$ & \textsc{S2m}$_{A,B}^{G}$ & \textsc{S2m}$_{G}^{A,B}$ \\
     \midrule
    \textsc{Sm}$_{A,B}^{G}$ & 0.0   & 1.5& 17.6$^\dagger$ & 19.0$^\dagger$ & 4.0 & 4.1 \\
    \textsc{Sm}$_{G}^{A,B}$ & - & 0.0 & 17.9$^\dagger$ & 19.5$^\dagger$& 3.9 & 4.0 \\
    \midrule
    \textsc{SB}$_{A,B}^{G}$ & - &-& 0.0 &8.1$^\dagger$ & 18.4$^\dagger$& 19.2$^\dagger$ \\
    \textsc{SB}$_{G}^{A,B}$ & - &-&  - & 0.0& 19.1$^\dagger$& 19.3$^\dagger$\\
    
    \midrule
    \textsc{S2m}$_{A,B}^{G}$ & -&-&-&- & 0.0 & 1.2 \\
    \textsc{S2m}$_{G}^{A,B}$ & -&-&-&-& & 0.0 \\
    \bottomrule
\end{tabular}}
\caption{Cross-metric comparison on individual graph rankings. 
\% of cases where metrics differ in their preference for one parse over the other. metric$_{X}^{Y}$: short for metric($X,Y$). $\dagger$ indicates significance in score differences assigned to parse pairs at p$<$0.005.
}
    \label{tab:spearman}
    \vspace{-3mm}
\end{table}

Table \ref{tab:spearman} shows that \textsc{Smatch} and \textsc{S$^2$match} both differ (significantly) from \textsc{SemBleu} in 15\% -- 20\% of cases. \textsc{Smatch} and \textsc{S$^2$match} differ on individual rankings in appr.\ 4\% of cases. Furthermore, we note a considerable amount of cases (8.1\%) where \textsc{SemBleu} disagrees with itself in the preference for one parse over the other.\footnote{i.e., \textsc{SB}(A,G)$>$\textsc{SB}(B,G) albeit \textsc{SB}(G,A)$<$\textsc{SB}(G,B).}

\begin{figure}[ht!]
\centering
\begin{subfigure}{.48\textwidth}
\begin{ssmall}
\begin{Verbatim}[commandchars=\\\{\},codes={\catcode`$=3\catcode`^=7}]
\textbf{------------Gold Graph & Input Sentence-------------}

\textbf{(t / thing :quant 2}            \textit{\textbf{"Legally, there are}}
\textbf{  :ARG2-of (r / remedy-01)}     \textit{\textbf{two remedies."}}
\textbf{  :mod \colorbox{lavender(web)}{(l / law))}}

\textbf{----------CAMR Parse------------JAMR Parse----------}

\textbf{(x6 / remedy-01         \colorbox{lavender(web)}{(l / legally}}
\textbf{  :quant 2)               :manner-of (r / remedy-01}
\textbf{                          :quant 2))}

\textbf{------------Alignments (parse, gold)----------------}

\textbf{\textsc{Smatch}:}  x6=r                l=NULL, r=r,
\textbf{\textsc{S$^2$match}:} x6=r                \colorbox{lavender(web)}{l=l}, r=r

\textbf{---------------------Scores-------------------------}

\textbf{\textsc{Smatch}:     0.200       \colorbox{lemonchiffon}{>>}      0.167}
\textbf{\textsc{S$^2$match}:    0.200       \colorbox{lemonchiffon}{<<}      0.252}
\textbf{----------------------------------------------------}
   \end{Verbatim}
   \end{ssmall}
   \vspace{-3mm}
\caption{$\textsc{S$^2$match}$ prefers JAMR parse.}
\label{fig:changedranking1}
\end{subfigure}
\begin{subfigure}{.48\textwidth}
\begin{ssmall}
\begin{Verbatim}[commandchars=\\\{\},codes={\catcode`$=3\catcode`^=7}]

\textbf{------------Gold Graph & Input Sentence-------------}

\textbf{\colorbox{lavender(web)}{(n3 / navy} }          \textit{\textbf{"The Navy of the Russian}}
\textbf{  :mod (c / country}    \textit{\textbf{Federation is in poor shape."}}
\textbf{    :name (n2 / name }
\textbf{     :op1 "Russian" }
\textbf{     :op2 "Federation"))}
\textbf{  :mod (s / shape}
\textbf{    :mod \colorbox{lavender(web)}{(p / poverty)))}}
           
\textbf{----------CAMR Parse------------JAMR Parse----------}

\textbf{\colorbox{lavender(web)}{(x2 / military}          (s / shape-01}
\textbf{  :name (n / name         :ARG1 (c / country}
\textbf{    :op1 "Navy")            :name (n / name}
\textbf{  :poss (x5 / country         :op1 "Russian"}
\textbf{    :name (n1 / name          :op2 "Federation")}
\textbf{      :op1 "Russia"         :poss (o / organization}
\textbf{      :op2 "Federation"))     :name (n2 / name}
\textbf{  :prep-in (x10 / shape-01      :op1 "Navy" :op2 "of"}
\textbf{    :mod \colorbox{lavender(web)}{(x9 / poor)))}          :op3 "the")))}
\textbf{                            :manner (p / poor))}

\textbf{------------Alignments (parse, gold)----------------}

\textbf{\textsc{Smatch}}:  x2=NULL, n=NULL,     o=n3, n2=NULL,
         x5=c , n1=n2,        c=c, n=n2,  
         x10=n3,x x9=s        p=NULL, s=NULL                            
\textbf{\textsc{S$^2$match}}: \colorbox{lavender(web)}{x2=n3}, n=NULL,      o=n3, n2=NULL,
         x5=c , n1=n2,        c=c, n=n2,  
         x10=s,x \colorbox{lavender(web)}{x9=p}        p=p, s=s
    
\textbf{---------------------Scores-------------------------}

\textbf{\textsc{Smatch}:     0.357       \colorbox{lemonchiffon}{<<}      0.387}
\textbf{\textsc{S$^2$match}:    0.488       \colorbox{lemonchiffon}{>>}      0.460}
\textbf{----------------------------------------------------}
   \end{Verbatim}
   \end{ssmall}
\caption{$\textsc{S$^2$match}$ prefers CAMR parse.}
\label{fig:changedranking2}
\end{subfigure}
\caption{Two examples, where $\textsc{S$^2$match}$ disagrees with \textsc{Smatch} in its preference of a candidate parse (for clarity, wiki-links are omitted in this display).}
\label{fig:changedrankingmain}
\end{figure}

The differing preferences of \textsc{S$^{(2)}$match} for either candidate parse can be the outcome of small divergences due to the alignment search or because \textsc{S$^2$match} accounts for the lexical similarity of concepts, perhaps supported by a new variable alignment. Figure \ref{fig:changedrankingmain} shows two examples where \textsc{S$^2$match} prefers a different candidate parse compared to \textsc{Smatch}. In the first example (Figure \ref{fig:changedranking1}), \textsc{S$^2$match} prefers the parse produced by JAMR and changes the alignment \textit{legally-NULL} (\textsc{Smatch}) to \textit{legally-law} (\textsc{S$^2$match}). In the second example (\ref{fig:changedranking2}), \textsc{S$^2$match} prefers the parse produced by CAMR, because it detects the similarity between \textit{military} and \textit{navy} and \textit{poor} and \textit{poverty}. Therefore, \textsc{S$^2$match} can assess that the CAMR parse and the gold graph substantially agree on the root concept of the graph, which is not the case in the JAMR parse.

\paragraph{Quantitative study: parser rankings} Having seen that our metrics disagree on the ran\-king of individual graphs, we now 
quantify the effects on the ranking of parsers. 
We collect outputs of three state-of-art parsers on the test set of LDC2017T10: GPLA, a sequence-to-graph transducer (STOG) and a neural top-down parser (TOP-DOWN).
\begin{table}[]
    \centering
    \scalebox{0.62}{
    \begin{tabular}{@{}l@{~~}rrrr|rr@{}}
    \toprule
    &\multicolumn{4}{c}{metric scores}&\multicolumn{2}{c}{structure error}\\
    & \textsc{Sm} & \textsc{SB}$_{A}^{G}$ & \textsc{SB}$_{G}^{A}$ & \textsc{S$^2$m} & node degree &  graph density\\
    \midrule
        STOG 
        & 76.3$_{|1}$ & 59.6$_{|1}$& 58.9$_{|1}$ & 77.9$_{|1}$   & 0.08 & 0.0069 \\
        GPLA 
        & 74.5$_{|2}$& 54.2$_{|3}$ & 52.9$_{|3}$ & 76.2$_{|2}$  & 0.08 & 0.0068 \\
        TOP-DOWN 
        & 73.2$_{|3}$ & 54.5$_{|2}$&53.1$_{|2}$ & 75.0$_{|3}$ & 0.11 & 0.0078\\
        \bottomrule
    \end{tabular}}
    \caption{Ranking parsers:  STOG (\citeauthor{DBLP:journals/corr/zhang19}); GPLA (\citeauthor{lyu-titov-2018-amr}); TOP-DOWN \cite{cai-lam-2019-core}. The structure error is defined as $\frac{1}{1371}\sum_{i=1}^{1371} | f(gold_i) - f(pred_i) |$, where $f$ either is node degree or graph density. All four metrics differ significantly 
    in their scores (paired t-test, p$<$0.05).
    }
    \label{tab:parsercomp}
    \vspace{-3mm}
\end{table}

Table \ref{tab:parsercomp} shows that \textsc{Smatch} and \textsc{S$^2$match} agree on the ranking of all three parsers, but both 
disagree with \textsc{SemBleu} on the ranks of the 2$^{nd}$ and 3$^{rd}$ parser: unlike \textsc{SemBleu}, the \textsc{Smatch} variants rate GPLA 
higher
than TOP-DOWN.
A factor that may have contributed to the different rankings perhaps lies in \textsc{SemBleu}'s biases towards connected nodes: Compared with TOP-DOWN, GPLA delivers more complex parses, with more edges (avg.\ $|E|$: 32.8 vs.\ 32.1) and higher graph density (avg.\ density: 0.065 vs.\ 0.059).
This is a nice property, since it indicates that the graphs 
of GPLA better resemble
the rich gold graph structures (avg.\ density: 0.063, avg.\ $|E|$: 34.2).  When inspecting this 
more closely, and looking at single (parse, gold) pairs, we observe further evidence for this: the structural error, in degree and density,
is lower for GPLA than for TOP-DOWN (Table \ref{tab:parsercomp}, right columns), with an error reduction of -27\% (degree, 0.08 vs.\ 0.11) and -14\% (density, 0.0067 vs.\ 0.0078). 

In sum, by building graphs of higher complexity, 
GPLA takes a greater risk when attaching wrong concepts to connected nodes where errors are penalized more strongly by \textsc{SemBleu} than \textsc{Smatch}, according
to the biases we have studied in \S \ref{sec:bias} (Table \ref{tab:scenarios}). In that sense, STOG also takes more risks, but it 
may get more of such concepts right and so the bias transitions from penalty to reward, potentially explaining the large performance $\Delta$ (+6) of STOG to the other parsers, as measured by \textsc{SemBleu}, in contrast to \textsc{S(2)Match} ($\Delta$: +2).

\section{Summary of our metric analyses}\label{subsec:summ}

\begin{table}
    \centering\scalebox{0.67}{
    \begin{tabular}{@{}lllll@{}}
    \toprule
       principle & \textsc{Smatch} & \textsc{SemBleu} & \textsc{S$^2$match} & Sec.\\
       \midrule
        I. Cont., non-neg.\  \&& &  &&  \\
        ~~~upper-bound & \cmark & \cmark &\cmark& - \\
        II. id.\ of indescernibles\ & \cmark$_\epsilon$ & \xmark & \cmark$_{\delta < \epsilon}$& \S \ref{par:id} \\
        III. symmetry  & \cmark$_\epsilon$ & \xmark & \cmark$_{\delta < \epsilon}$& \S \ref{par:symm}\\
        IV. determinacy & \cmark$_\epsilon$ & \cmark & \cmark$_{\delta < \epsilon}$ &\S \ref{par:det}\\
        V. low bias& \cmark & \xmark & \cmark &\S \ref{sec:bias}\\
        VI. symbolic graph matching & \cmark & \xmark &\cmark& \S \ref{subsec:alignornot} \\
        VII. graded graph matching & \xmark & \xmark & \cmark$^{LEX}$&\S \ref{sec:s2} \\
        \bottomrule
    \end{tabular}}
    \caption{Evaluation of three AMR metrics using our seven
    principles. 
    \cmark$_\epsilon$:  fulfilled with a very small $\epsilon$-error.
    }
    \vspace{-3mm}
    \label{tab:result}
\end{table}

Table \ref{tab:result} sum\-ma\-ri\-zes our analyses' integral results. Principle I is fulfilled by all metrics as they exhibit \textit{continuity, non-negativity and an upper bound}. Principle II, however, is not satisfied by \textsc{SemBleu} since it can mistake two graphs of different meaning as equivalent. This is because it ablates a variable-alignment and therefore cannot capture facets of coreference. Yet, a positive outcome of this is that it is \textit{fast to compute}. This could make it first choice in some recent AMR parsing approaches that use reinforcement learning \cite{DBLP:journals/corr/abs-1905-13370}, where rapid feedback is needed. 
It also marks a point by fully satisfying Principle IV,  yielding fully deterministic results. \textsc{Smatch}, by contrast, either needs to resort to a costly ILP solution or (in practice) uses 
hill-climbing 
with multiple restarts to reduce 
divergence to a negligible amount.

A central insight brought out by our ana\-ly\-sis is that
\textsc{SemBleu} exhibits \textit{biases} that are hard to control.
This is cau\-sed by two (interacting) factors: (i) The extraction of k-grams is applied on the graph top to bottom and 
visits some nodes more frequently than oth\-ers. (ii) 
It raises some (but not all) leaf nodes to connected nodes, and 
these nodes will be overly frequently contained in 
extracted k-grams. We have shown that these two factors in com\-bi\-na\-tion lead to large biases that researchers 
should
be aware of when using \textsc{SemBleu} 
(\S \ref{sec:bias}). Its ancestor \textsc{Bleu} does not suffer from such biases since
it extracts k-grams linearly from a sentence. 

Given that \textsc{SemBleu} is built on \textsc{Bleu}, it is inherently \textit{asymmetric}. However, 
we have shown that 
the a\-sym\-me\-try (Principle III) measured for \textsc{Bleu} in MT is am\-pli\-fied by \textsc{SemBleu} in AMR, mainly due to the biases it incurs (Principle V). While asymmetry can be tolerated in parser evaluation if outputs are compared against gold 
graphs in a standardized manner, it is difficult to apply an asymmetric metric to measure IAA or to compare parses for detecting paraphrases, or in tri-parsing, where no reference is available. If the asymmetry is amplified by a bias, it becomes harder to judge the scores.
Finally,
considering that \textsc{SemBleu} does not match AMR graphs on the graph-level but matches extracted bags-of-k-grams, 
it turns out that it cannot be categorized as a graph matching algorithm as defined in Principle VI.

Principle VI is fulfilled by \textsc{Smatch} without
any transformation on 
AMR graphs. 
It searches for an optimal variable align\-ment and 
counts 
matching 
triples. As a corollary, 
it fulfills principles I, II, III and V. The fact that \textsc{Smatch} fulfills all but one principle backs up many 
prior works that use it as 
sole criterion for IAA and parse evaluation.

Our principles also 
helped us detect a weakness 
of all present AMR metrics: 
they operate on a discrete level and cannot assess graded meaning
differences.
As a first step, 
we propose
\textsc{S$^2$match}: it preserves beneficial properties of \textsc{Smatch} but is 
benevolent to
slight lexical meaning deviations. Besides parser evaluation, this property makes the metric also more suitable for other tasks, e.g., it can be used as a kernel in an SVM that classifies AMRs to determine whether two sentences are equivalent in meaning. In such a case, \textsc{S$^2$match} is bound to detect meaning-similarities that cannot be captured by \textsc{Smatch} or \textsc{SemBleu}, e.g., due to paraphrases being projected into the parses.

\section{Related work}

Developing
similarity metrics for meaning representations (MRs) is important, as it, i.a., affects 
semantic parser evaluation and 
computation of IAA statistics
for sembanking. MRs 
are de\-sig\-ned to re\-pre\-sent the meaning of text  in a 
well-defined, interpretable form that is able to 
identify meaning differences and support in\-fe\-rence. \citet{bos-first-order,bos2019separating} has shown how AMR can be translated to FOL, a well-established MR for\-ma\-lism.
Discourse Representation Theory (DRT, 
\citet{Kamp81,KampReyle:93}) 
is based on and extends FOL to
 discourse representation. A recent shared task on DRS parsing used the \textsc{Counter} metric  \cite{abzianidze2019first,evang2019transition}, an adaption of \textsc{Smatch}, underlining \textsc{Smatch}'s general applicability. 
Its extension \textsc{S$^2$match} may also prove beneficial for DRS. 

Other research into AMR metrics aims at making the comparison fairer by 
normalizing
graphs \cite{goodman2019amr}. \citet{anchieta2019sema} argue that one should not, e.g., insert an extra \textit{is-root} node when comparing AMR graphs (as done in \textsc{SemBleu} and \textsc{Smatch}). 
\citet{DBLP:journals/corr/DamonteCS16} extend 
\textsc{Smatch} 
to analyze individual 
AMR facets 
(co-reference, WSD, etc.). \citet{cai-lam-2019-core} adapt \textsc{Smatch} to analyze their parser's performance in predicting triples that are in close proximity to the root. Our metric \textsc{S$^2$match} allows for straightforward integration of these approaches.

\paragraph{Computational AMR tasks}
Since the introduction of AMR, 
 many 
 AMR-related tasks have emer\-ged. Most prominent is AMR parsing
 \cite{wang-etal-2015-boosting,wang2016camr,DBLP:journals/corr/DamonteCS16,konstas2017neural,lyu-titov-2018-amr,DBLP:journals/corr/zhang19}. The inverse task 
 generates text from AMR graphs \cite{song-etal-2017-amr,DBLP:journals/corr/song18,DBLP:journals/corr/damonte19}. \citet{opitz-frank-2019-automatic} rate the quality of automatic
 AMR parses
 without costly gold data.

\section{Conclusion}

We motivated seven principles for metrics measuring the similarity of graph-based (Ab\-stract) Meaning Representations, from mathematical, linguistic and engineering perspectives. 
A metric that ful\-fills all
principles 
is applicable to a wide spectrum of cases, ranging from parser evaluation to sound IAA calculation. Hence \textbf{(i) our prin\-ci\-ples can inform (A)MR researchers who desire to com\-pare and select among metrics}, and
\textbf{(ii) 
they 
ease and guide the development of new metrics}.

We provided examples for both scenarios. 
We
show\-cased (i)
by utilizing our principles as guide\-li\-nes for an in-depth analysis of two AMR metrics: 
\textsc{Smatch} and the recent \textsc{SemBleu} metrics, two quite distinct approaches. 
Our analysis
uncovered that the latter 
does not satisfy some principles, which 
might
reduce its safety and applicability. In line of (ii), we target the fulfilment of all seven principles and propose \textsc{S$^2$match}, a metric that accounts for graded similarity of concepts 
as atomic graph components. In future work, we 
aim for a metric that 
accounts for graded compositional similarity of subgraphs.

\subsubsection*{Acknowledgements.} We are grateful to the anonymous reviewers and the action editors for their valuable time and comments. This work has been partially funded by DFG within the project \textit{ExpLAIN. Between the Lines – Knowledge-based Analysis of Argumentation in a Formal Argumentation Inference System}, FR 1707/-4-1, as part of the RATIO Priority Program and by the
the \textit{Leibniz ScienceCampus Empirical Linguistics \& Computational Language Modeling}, supported by Leibniz Association grant no.\ SAS2015-IDS-LWC and by the Ministry of Science, Research, and Art of Baden-Wurttemberg.

\bibliography{main.bib}

\begin{thebibliography}{41}
\expandafter\ifx\csname natexlab\endcsname\relax\def\natexlab#1{#1}\fi

\bibitem[{Abzianidze et~al.(2019)Abzianidze, van Noord, Haagsma, and
  Bos}]{abzianidze2019first}
Lasha Abzianidze, Rik van Noord, Hessel Haagsma, and Johan Bos. 2019.
\newblock \href {https://doi.org/10.18653/v1/W19-1201} {{The First Shared Task
  on Discourse Representation Structure Parsing}}.
\newblock In \emph{Proceedings of the {IWCS} Shared Task on Semantic Parsing},
  Gothenburg, Sweden.

\bibitem[{Anchi\^{e}ta et~al.(2019)Anchi\^{e}ta, Cabezudo, and
  Pardo}]{anchieta2019sema}
Rafael~Torres Anchi\^{e}ta, Marco Antonio~Sobrevilla Cabezudo, and Thiago
  Alexandre~Salgueiro Pardo. 2019.
\newblock {SEMA: an Extended Semantic Evaluation for AMR}.
\newblock In \emph{(To appear) Proceedings of the 20th Computational
  Linguistics and Intelligent Text Processing}. Springer International
  Publishg.

\bibitem[{Banarescu et~al.(2013)Banarescu, Bonial, Cai, Georgescu, Griffitt,
  Hermjakob, Knight, Koehn, Palmer, and Schneider}]{banarescu2013abstract}
Laura Banarescu, Claire Bonial, Shu Cai, Madalina Georgescu, Kira Griffitt, Ulf
  Hermjakob, Kevin Knight, Philipp Koehn, Martha Palmer, and Nathan Schneider.
  2013.
\newblock {Abstract meaning representation for sembanking}.
\newblock In \emph{Proceedings of the 7th Linguistic Annotation Workshop and
  Interoperability with Discourse}, pages 178--186.

\bibitem[{Bos(2016)}]{bos-first-order}
Johan Bos. 2016.
\newblock \href {https://doi.org/10.1162/COLI\_a\_00257} {{Expressive Power of
  Abstract Meaning Representations}}.
\newblock \emph{Computational Linguistics}, 42(3):527--535.

\bibitem[{Bos(2019)}]{bos2019separating}
Johan Bos. 2019.
\newblock {Separating Argument Structure from Logical Structure in AMR}.
\newblock \emph{arXiv preprint arXiv:1908.01355}.

\bibitem[{Budanitsky and Hirst(2006)}]{budanitsky-hirst-2006-evaluating}
Alexander Budanitsky and Graeme Hirst. 2006.
\newblock \href {https://doi.org/10.1162/coli.2006.32.1.13} {{Evaluating
  {W}ord{N}et-based Measures of Lexical Semantic Relatedness}}.
\newblock \emph{Computational Linguistics}, 32(1):13--47.

\bibitem[{Cai and Lam(2019)}]{cai-lam-2019-core}
Deng Cai and Wai Lam. 2019.
\newblock \href {https://doi.org/10.18653/v1/D19-1393} {{Core Semantic First: A
  Top-down Approach for {AMR} Parsing}}.
\newblock In \emph{Proceedings of the 2019 Conference on Empirical Methods in
  Natural Language Processing and the 9th International Joint Conference on
  Natural Language Processing (EMNLP-IJCNLP)}, pages 3797--3807, Hong Kong,
  China.

\bibitem[{Cai and Knight(2013)}]{cai-knight-2013-smatch}
Shu Cai and Kevin Knight. 2013.
\newblock \href {https://www.aclweb.org/anthology/P13-2131} {{{S}match: an
  Evaluation Metric for Semantic Feature Structures}}.
\newblock In \emph{Proceedings of the 51st Annual Meeting of the Association
  for Computational Linguistics (Volume 2: Short Papers)}, pages 748--752,
  Sofia, Bulgaria.

\bibitem[{Cer et~al.(2017)Cer, Diab, Agirre, Lopez-Gazpio, and
  Specia}]{cer-etal-2017-semeval}
Daniel Cer, Mona Diab, Eneko Agirre, I{\~n}igo Lopez-Gazpio, and Lucia Specia.
  2017.
\newblock \href {https://doi.org/10.18653/v1/S17-2001} {{SemEval-2017 Task 1:
  Semantic Textual Similarity Multilingual and Crosslingual Focused
  Evaluation}}.
\newblock In \emph{Proceedings of the 11th International Workshop on Semantic
  Evaluation ({S}em{E}val-2017)}, pages 1--14, Vancouver, Canada.

\bibitem[{Chen and Cherry(2014)}]{chen-cherry:2014:W14-33}
Boxing Chen and Colin Cherry. 2014.
\newblock \href {http://www.aclweb.org/anthology/W/W14/W14-3346} {{A Systematic
  Comparison of Smoothing Techniques for Sentence-Level BLEU}}.
\newblock In \emph{Proceedings of the Ninth Workshop on Statistical Machine
  Translation}, pages 362--367, Baltimore, Maryland, USA.

\bibitem[{Damonte and Cohen(2019)}]{DBLP:journals/corr/damonte19}
Marco Damonte and Shay~B. Cohen. 2019.
\newblock \href {https://doi.org/10.18653/v1/N19-1366} {{Structural Neural
  Encoders for {AMR}-to-text Generation}}.
\newblock In \emph{Proceedings of the 2019 Conference of the North {A}merican
  Chapter of the Association for Computational Linguistics: Human Language
  Technologies, Volume 1 (Long and Short Papers)}, pages 3649--3658,
  Minneapolis, Minnesota.

\bibitem[{Damonte et~al.(2017)Damonte, Cohen, and
  Satta}]{DBLP:journals/corr/DamonteCS16}
Marco Damonte, Shay~B. Cohen, and Giorgio Satta. 2017.
\newblock \href {https://www.aclweb.org/anthology/E17-1051} {{An Incremental
  Parser for Abstract Meaning Representation}}.
\newblock In \emph{Proceedings of the 15th Conference of the {E}uropean Chapter
  of the Association for Computational Linguistics: Volume 1, Long Papers},
  pages 536--546, Valencia, Spain.

\bibitem[{Edmonds and Hirst(2002)}]{Edmonds:2002:NLC:643081.643082}
Philip Edmonds and Graeme Hirst. 2002.
\newblock \href {https://doi.org/10.1162/089120102760173625} {{Near-synonymy
  and Lexical Choice}}.
\newblock \emph{Computational Linguistics}, 28(2):105--144.

\bibitem[{Evang(2019)}]{evang2019transition}
Kilian Evang. 2019.
\newblock \href {https://doi.org/10.18653/v1/W19-1202} {{Transition-based {DRS}
  Parsing Using Stack-{LSTM}s}}.
\newblock In \emph{Proceedings of the {IWCS} Shared Task on Semantic Parsing},
  Gothenburg, Sweden.

\bibitem[{Flanigan et~al.(2014)Flanigan, Thomson, Carbonell, Dyer, and
  Smith}]{flanigan-etal-2014-discriminative}
Jeffrey Flanigan, Sam Thomson, Jaime Carbonell, Chris Dyer, and Noah~A. Smith.
  2014.
\newblock \href {https://doi.org/10.3115/v1/P14-1134} {{A Discriminative
  Graph-Based Parser for the Abstract Meaning Representation}}.
\newblock In \emph{Proceedings of the 52nd Annual Meeting of the Association
  for Computational Linguistics (Volume 1: Long Papers)}, pages 1426--1436,
  Baltimore, Maryland.

\bibitem[{Goodman(2019)}]{goodman2019amr}
Michael~Wayne Goodman. 2019.
\newblock {AMR} normalization for fairer evaluation.
\newblock In \emph{Proceedings of the 33rd Pacific Asia Conference on Language,
  Information, and Computation}, Hakodate.

\bibitem[{Huck et~al.(2017)Huck, Braune, and Fraser}]{huck-etal-2017-lmu}
Matthias Huck, Fabienne Braune, and Alexander Fraser. 2017.
\newblock \href {https://doi.org/10.18653/v1/W17-4730} {{{LMU} Munich{'}s
  Neural Machine Translation Systems for News Articles and Health Information
  Texts}}.
\newblock In \emph{Proceedings of the Second Conference on Machine
  Translation}, pages 315--322, Copenhagen, Denmark.

\bibitem[{Inkpen and Hirst(2006)}]{Inkpen:2006:BUL:1169207.1169210}
Diana Inkpen and Graeme Hirst. 2006.
\newblock \href {https://doi.org/10.1162/coli.2006.32.2.223} {{Building and
  Using a Lexical Knowledge Base of Near-Synonym Differences}}.
\newblock \emph{Computational Linguistics}, 32(2):223--262.

\bibitem[{Jaccard(1912)}]{jaccard1912distribution}
Paul Jaccard. 1912.
\newblock {The distribution of the flora in the alpine zone}.
\newblock \emph{New phytologist}, 11(2):37--50.

\bibitem[{Kamp(1981)}]{Kamp81}
Hans Kamp. 1981.
\newblock {A theory of truth and semantic representation}.
\newblock \emph{Formal semantics: The Essential Readings}, pages 189--222.

\bibitem[{Kamp and Reyle(1993)}]{KampReyle:93}
Hans Kamp and Uwe Reyle. 1993.
\newblock \emph{{From Discourse to Logic. Introduction to Model-theoretic
  Semantics of Natural Language, Formal Logic and Discourse Representation
  Theory}}.
\newblock Kluwer, Dordrecht.

\bibitem[{Konstas et~al.(2017)Konstas, Iyer, Yatskar, Choi, and
  Zettlemoyer}]{konstas2017neural}
Ioannis Konstas, Srinivasan Iyer, Mark Yatskar, Yejin Choi, and Luke
  Zettlemoyer. 2017.
\newblock \href {https://doi.org/10.18653/v1/P17-1014} {{Neural {AMR:}
  Sequence-to-Sequence Models for Parsing and Generation}}.
\newblock In \emph{Proceedings of the 55th Annual Meeting of the Association
  for Computational Linguistics, {ACL} 2017, Vancouver, Canada, July 30 -
  August 4, Volume 1: Long Papers}, pages 146--157.

\bibitem[{Lyu and Titov(2018)}]{lyu-titov-2018-amr}
Chunchuan Lyu and Ivan Titov. 2018.
\newblock \href {https://www.aclweb.org/anthology/P18-1037} {{{AMR} Parsing as
  Graph Prediction with Latent Alignment}}.
\newblock In \emph{Proceedings of the 56th Annual Meeting of the Association
  for Computational Linguistics (Volume 1: Long Papers)}, pages 397--407,
  Melbourne, Australia.

\bibitem[{Ma et~al.(2018)Ma, Bojar, and Graham}]{DBLP:conf/wmt/MaBG18}
Qingsong Ma, Ondrej Bojar, and Yvette Graham. 2018.
\newblock \href {https://www.aclweb.org/anthology/W18-6450/} {{Results of the
  {WMT18} Metrics Shared Task: Both characters and embeddings achieve good
  performance}}.
\newblock In \emph{Proceedings of the Third Conference on Machine Translation:
  Shared Task Papers, {WMT} 2018, Belgium, Brussels, October 31 - November 1,
  2018}, pages 671--688.

\bibitem[{Marelli et~al.(2014)Marelli, Bentivogli, Baroni, Bernardi, Menini,
  and Zamparelli}]{marelli-etal-2014-semeval}
Marco Marelli, Luisa Bentivogli, Marco Baroni, Raffaella Bernardi, Stefano
  Menini, and Roberto Zamparelli. 2014.
\newblock \href {https://doi.org/10.3115/v1/S14-2001} {{{S}em{E}val-2014 Task
  1: Evaluation of Compositional Distributional Semantic Models on Full
  Sentences through Semantic Relatedness and Textual Entailment}}.
\newblock In \emph{Proceedings of the 8th International Workshop on Semantic
  Evaluation ({S}em{E}val 2014)}, pages 1--8, Dublin, Ireland.

\bibitem[{Naseem et~al.(2019)Naseem, Shah, Wan, Florian, Roukos, and
  Ballesteros}]{DBLP:journals/corr/abs-1905-13370}
Tahira Naseem, Abhishek Shah, Hui Wan, Radu Florian, Salim Roukos, and Miguel
  Ballesteros. 2019.
\newblock \href {https://doi.org/10.18653/v1/P19-1451} {{Rewarding Smatch:
  Transition-Based AMR Parsing with Reinforcement Learning}}.
\newblock In \emph{Proceedings of the 57th Annual Meeting of the Association
  for Computational Linguistics}, pages 4586--4592, Florence, Italy.

\bibitem[{Opitz and Frank(2019)}]{opitz-frank-2019-automatic}
Juri Opitz and Anette Frank. 2019.
\newblock \href {https://www.aclweb.org/anthology/S19-1024} {{Automatic
  Accuracy Prediction for {AMR} Parsing}}.
\newblock In \emph{Proceedings of the Eighth Joint Conference on Lexical and
  Computational Semantics (*{SEM} 2019)}, pages 212--223, Minneapolis,
  Minnesota.

\bibitem[{Papadimitriou et~al.(2010)Papadimitriou, Dasdan, and
  Garcia-Molina}]{papadimitriou2010web}
Panagiotis Papadimitriou, Ali Dasdan, and Hector Garcia-Molina. 2010.
\newblock {Web graph similarity for anomaly detection}.
\newblock \emph{Journal of Internet Services and Applications}, 1(1):19--30.

\bibitem[{Papineni et~al.(2002)Papineni, Roukos, Ward, and
  Zhu}]{papineni2002bleu}
Kishore Papineni, Salim Roukos, Todd Ward, and Wei-Jing Zhu. 2002.
\newblock \href {https://doi.org/10.3115/1073083.1073135} {{{B}leu: a Method
  for Automatic Evaluation of Machine Translation}}.
\newblock In \emph{Proceedings of the 40th Annual Meeting of the Association
  for Computational Linguistics}, pages 311--318, Philadelphia, Pennsylvania,
  USA.

\bibitem[{Pennington et~al.(2014)Pennington, Socher, and
  Manning}]{pennington2014glove}
Jeffrey Pennington, Richard Socher, and Christopher Manning. 2014.
\newblock \href {https://doi.org/10.3115/v1/D14-1162} {{{G}lo{V}e: Global
  Vectors for Word Representation}}.
\newblock In \emph{Proceedings of the 2014 Conference on Empirical Methods in
  Natural Language Processing ({EMNLP})}, pages 1532--1543, Doha, Qatar.

\bibitem[{Real and Vargas(1996)}]{real1996probabilistic}
Raimundo Real and Juan~M Vargas. 1996.
\newblock {The probabilistic basis of Jaccard's index of similarity}.
\newblock \emph{Systematic biology}, 45(3):380--385.

\bibitem[{Riesen et~al.(2010)Riesen, Jiang, and Bunke}]{riesen2010exact}
Kaspar Riesen, Xiaoyi Jiang, and Horst Bunke. 2010.
\newblock {Exact and inexact graph matching: Methodology and applications}.
\newblock In \emph{Managing and Mining Graph Data}, pages 217--247. Springer.

\bibitem[{Schenker et~al.(2005)Schenker, Bunke, Last, and
  Kandel}]{10.5555/1121730}
Adam Schenker, Horst Bunke, Mark Last, and Abraham Kandel. 2005.
\newblock \emph{{Graph-Theoretic Techniques for Web Content Mining}}.
\newblock World Scientific Publishing Co., Inc., USA.

\bibitem[{Song and Gildea(2019)}]{DBLP:journals/corr/abs-1905-10726}
Linfeng Song and Daniel Gildea. 2019.
\newblock \href {https://doi.org/10.18653/v1/P19-1446} {{{S}em{B}leu: A Robust
  Metric for {AMR} Parsing Evaluation}}.
\newblock In \emph{Proceedings of the 57th Annual Meeting of the Association
  for Computational Linguistics}, pages 4547--4552, Florence, Italy.

\bibitem[{Song et~al.(2017)Song, Peng, Zhang, Wang, and
  Gildea}]{song-etal-2017-amr}
Linfeng Song, Xiaochang Peng, Yue Zhang, Zhiguo Wang, and Daniel Gildea. 2017.
\newblock \href {https://doi.org/10.18653/v1/P17-2002} {{{AMR}-to-text
  Generation with Synchronous Node Replacement Grammar}}.
\newblock In \emph{Proceedings of the 55th Annual Meeting of the Association
  for Computational Linguistics (Volume 2: Short Papers)}, pages 7--13,
  Vancouver, Canada.

\bibitem[{Song et~al.(2016)Song, Zhang, Peng, Wang, and
  Gildea}]{DBLP:journals/corr/SongZPWG16}
Linfeng Song, Yue Zhang, Xiaochang Peng, Zhiguo Wang, and Daniel Gildea. 2016.
\newblock \href {https://doi.org/10.18653/v1/D16-1224} {{{AMR}-to-text
  generation as a Traveling Salesman Problem}}.
\newblock In \emph{Proceedings of the 2016 Conference on Empirical Methods in
  Natural Language Processing}, pages 2084--2089, Austin, Texas.

\bibitem[{Song et~al.(2018)Song, Zhang, Wang, and
  Gildea}]{DBLP:journals/corr/song18}
Linfeng Song, Yue Zhang, Zhiguo Wang, and Daniel Gildea. 2018.
\newblock \href {https://doi.org/10.18653/v1/P18-1150} {{A Graph-to-Sequence
  Model for {AMR}-to-Text Generation}}.
\newblock In \emph{Proceedings of the 56th Annual Meeting of the Association
  for Computational Linguistics (Volume 1: Long Papers)}, pages 1616--1626,
  Melbourne, Australia.

\bibitem[{Wang et~al.(2016)Wang, Pradhan, Pan, Ji, and Xue}]{wang2016camr}
Chuan Wang, Sameer Pradhan, Xiaoman Pan, Heng Ji, and Nianwen Xue. 2016.
\newblock \href {https://doi.org/10.18653/v1/S16-1181} {{{CAMR} at
  {S}em{E}val-2016 Task 8: An Extended Transition-based {AMR} Parser}}.
\newblock In \emph{Proceedings of the 10th International Workshop on Semantic
  Evaluation ({S}em{E}val-2016)}, pages 1173--1178, San Diego, California.

\bibitem[{Wang et~al.(2015)Wang, Xue, and Pradhan}]{wang-etal-2015-boosting}
Chuan Wang, Nianwen Xue, and Sameer Pradhan. 2015.
\newblock \href {https://doi.org/10.3115/v1/P15-2141} {{Boosting
  Transition-based {AMR} Parsing with Refined Actions and Auxiliary
  Analyzers}}.
\newblock In \emph{Proceedings of the 53rd Annual Meeting of the Association
  for Computational Linguistics and the 7th International Joint Conference on
  Natural Language Processing (Volume 2: Short Papers)}, pages 857--862,
  Beijing, China.

\bibitem[{Yan et~al.(2016)Yan, Yin, Lin, Deng, Zha, and Yang}]{yan2016short}
Junchi Yan, Xu-Cheng Yin, Weiyao Lin, Cheng Deng, Hongyuan Zha, and Xiaokang
  Yang. 2016.
\newblock {A short survey of recent advances in graph matching}.
\newblock In \emph{Proceedings of the 2016 ACM on International Conference on
  Multimedia Retrieval}, pages 167--174. ACM.

\bibitem[{Zhang et~al.(2019)Zhang, Ma, Duh, and
  Van~Durme}]{DBLP:journals/corr/zhang19}
Sheng Zhang, Xutai Ma, Kevin Duh, and Benjamin Van~Durme. 2019.
\newblock \href {https://doi.org/10.18653/v1/P19-1009} {{{AMR} Parsing as
  Sequence-to-Graph Transduction}}.
\newblock In \emph{Proceedings of the 57th Annual Meeting of the Association
  for Computational Linguistics}, pages 80--94, Florence, Italy.

\end{thebibliography}
\bibliographystyle{acl_natbib}
\end{document}